\documentclass[sigconf]{acmart}
\usepackage[shortlabels]{enumitem}
\usepackage{array}
\usepackage{makecell}

\usepackage{pifont}
\usepackage{graphicx}
\usepackage{caption}
\usepackage{subcaption}
\usepackage{xcolor}
\usepackage{soul}
\usepackage{booktabs}
\usepackage{listings}
\usepackage{csquotes}
\usepackage{soul} % for \hl
\usepackage{xcolor} % for \sethlcolor
\usepackage[utf8]{inputenc}

\AtBeginDocument{%
  }

\setcopyright{acmlicensed}
\copyrightyear{2025}
\acmYear{2025}
\setcopyright{acmlicensed}\acmConference[ICAIF '25]{6th ACM International Conference on AI in Finance}{November 15--18, 2025}{Singapore, Singapore}
\acmBooktitle{6th ACM International Conference on AI in Finance (ICAIF '25), November 15--18, 2025, Singapore, Singapore}
\acmDOI{10.1145/3768292.3770385}
\acmISBN{979-8-4007-2220-2/2025/11}

\begin{document}

%% temporaneo, poi vediamo
\title{Prompting for Policy: Forecasting Macroeconomic
Scenarios with Synthetic LLM Personas}

\author{Giulia Iadisernia}
\authornote{Work done during an internship at Banca d’Italia. The views and opinions expressed are those of the authors and do not necessarily reflect the official policy or position of Banca d’Italia.}
\affiliation{%
  \institution{Banca d'Italia}
  \city{Rome}
  %\state{RM}
  \country{Italy}
}

\author{Carolina Camassa}
\authornote{Corresponding author.}
\affiliation{%
  \institution{Banca d'Italia}
  \city{Rome}
 % \state{RM}
  \country{Italy}
}

\email{carolina.camassa@bancaditalia.it}

\renewcommand{\shortauthors}{Iadisernia et al.}

\begin{abstract}
We evaluate whether persona-based prompting improves Large Language Model (LLM) performance on macroeconomic forecasting tasks. Using 2,368 economics-related personas from the PersonaHub corpus, we prompt GPT-4o to replicate the ECB Survey of Professional Forecasters across 50 quarterly rounds (2013-2025). We compare the persona-prompted forecasts against the human experts panel, across four target variables (HICP, core HICP, GDP growth, unemployment) and four forecast horizons. We also compare the results against 100 baseline forecasts without persona descriptions to isolate its effect. We report two main findings. Firstly, GPT-4o and human forecasters achieve remarkably similar accuracy levels, with differences that are statistically significant yet practically modest. Our out-of-sample evaluation on 2024-2025 data demonstrates that GPT-4o can maintain competitive forecasting performance on unseen events, though with notable differences compared to the in-sample period. Secondly, our ablation experiment reveals no measurable forecasting advantage from persona descriptions, suggesting these prompt components can be omitted to reduce computational costs without sacrificing accuracy. Our results provide evidence that GPT-4o can achieve competitive forecasting accuracy even on out-of-sample macroeconomic events, if provided with relevant context data, while revealing that diverse prompts produce remarkably homogeneous forecasts compared to human panels.
%This suggests that computational resources may be better allocated to model improvements rather than elaborate prompt engineering given the negligible impact of persona descriptions on forecasting performance.
\end{abstract}



\begin{teaserfigure}
  \includegraphics[width=\textwidth]{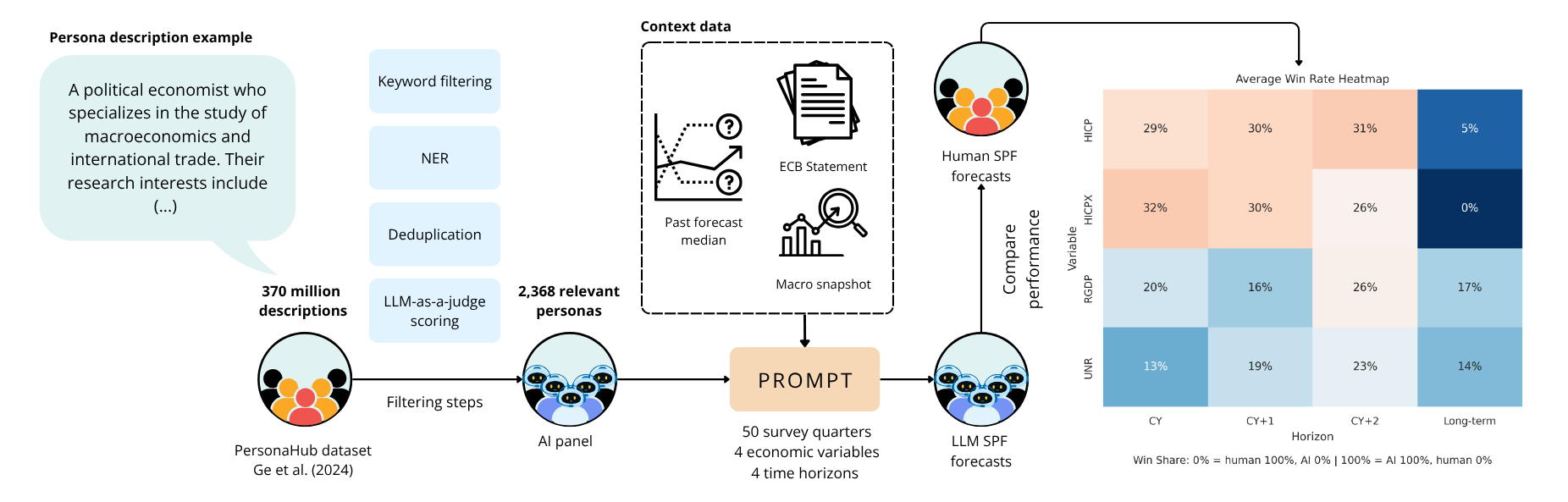}
  \caption{Experimental pipeline for evaluating synthetic LLM personas in macroeconomic forecasting. Starting from the PersonaHub corpus of 370M domain expert descriptions, we apply multi-stage filtering to extract 2,368 relevant personas. Each persona is then prompted to simulate responses to the ECB Survey of Professional Forecasters across 50 quarterly rounds (2013-2025), generating forecasts for four key macroeconomic variables (HICP inflation, core HICP, real GDP growth, unemployment) at multiple forecast horizons. The resulting 118,400 AI-generated forecasts are compared against human expert predictions to evaluate forecasting accuracy and the contribution of persona prompting to LLM performance in economic tasks.}
  \Description{Enjoying the baseball game from the third-base
  seats. Ichiro Suzuki preparing to bat.}
  \label{fig:teaser}
\end{teaserfigure}


\maketitle

\section{Introduction}

\begin{table*}[t]
\centering
\caption{The ECB Survey of Professional Forecasters: data description}
\begin{minipage}[t]{0.45\textwidth}
\centering
\subcaption{Main macroeconomic variables included in the ECB's Survey of Professional Forecasters}
\begin{tabular}{ll}
\toprule
\textbf{Variable} & \textbf{Definition} \\
\midrule
HICP & Harmonized Index of Consumer Prices (inflation)  \\
HICPX$^*$ &  Core HICP (excl. energy, food, alcohol, tobacco)\\
rGDP & Real GDP growth rate (annual \%) \\
UNR & Unemployment rate (\% of labor force)  \\ 
\bottomrule

% mancano solo i forecasts
\end{tabular}
\flushleft
{\footnotesize $^*$HICPX is included in SPF rounds since 2016Q4.}
\end{minipage}
\hfill
\begin{minipage}[t]{0.5\textwidth}
\centering
\subcaption{Dataset overview: number of human forecasts from ECB data VS simulated forecasts with AI}
\begin{tabular}{lrrr}
\toprule
\textbf{Data source} & \textbf{SPF Rounds} & \textbf{Forecasters} & \textbf{Forecasts} \\
\midrule
%70,080
%4,915
\multicolumn{4}{l}{\emph{Human SPF}} \\
\;In-sample & 44 & 56.2$^*$ & $\sim$2,473 \\
\;Out-of-sample & 6 & 58.4$^*$ & $\sim$350 \\[4pt]
\multicolumn{4}{l}{\emph{AI personas (per model)}} \\
\;With personas & 50 & 2,368 & 118,400 \\
\;No-persona baseline & 50 & 100 & 5,000 \\
\bottomrule
\end{tabular}
\end{minipage}
%\vspace{0.5em}
\flushright
\footnotesize
$^*$Average per round. 
\label{tab:data_overview}
\end{table*}
Macroeconomic forecasting has become increasingly critical for central bank communication and the transmission of monetary policy. The European Central Bank Survey of Professional Forecasters (ECB-SPF) \cite{spf-introduction}, conducted quarterly since 1999, represents one of the most systematic efforts to capture expert expectations of inflation, GDP growth, and unemployment in the euro area. Because these forecasts directly influence policy decisions and market expectations, producing accurate and consistent forecasts is essential for economic stability.

Large Language Models (LLMs) have emerged as promising tools for economic forecasting tasks, offering the potential to simulate expert judgment at scale. Yet, current applications face a key methodological limitation: most studies using LLMs to simulate economic forecasts \cite{hansen_simulating_2025} or model inflation expectations \cite{zarifhonarvar_evidence_2024} typically rely on one or few handcrafted \enquote{expert prompts}. 
Despite the increasing adoption of LLMs in economic research \cite{llm-econometric}, and the fact that LLM output is highly sensitive to prompt content and even formatting \cite{sclarquantifying}, there is a lack of empirical evidence on how prompt design affects performance in economics tasks. 

We build on this premise by producing the first systematic replication of the ECB Survey of Professional Forecasters using LLMs. Our main research question is: \textbf{Do sophisticated persona descriptions---detailed biographical prompts designed to simulate specific expert types---improve LLM performance in a macroeconomic forecasting task?}
%However, it's important to note that personas can have null or even adverse effects on performance, or due to the introduction of biases etc... \cite{kim2024persona}
We answer the question by extracting 2,368 economics-related synthetic biographies from the Persona Hub corpus\footnote{\href{https://huggingface.co/datasets/proj-persona/PersonaHub}{https://huggingface.co/datasets/proj-persona/PersonaHub}.}\cite{ge_scaling_2024}.
We evaluate the performance of these personas on 50 quarterly rounds (2013Q1-2025Q2) of the Survey of Professional Forecasters. This results in 118,400 AI-generated forecasts for four key macroeconomic variables at multiple forecast horizons. Our experimental design includes both \textit{in-sample} evaluation (2013Q1-2023Q4) and \textit{out-of-sample} testing on 2024-2025 data, which falls outside the model’s training data. 

Our study makes three main contributions. First, we conduct the first systematic replication of the ECB Survey of Professional Forecasters using LLMs, extending previous US-focused studies to European monetary policy contexts. Second, we implement a large-scale experimental design evaluating the macroeconomic forecasting performance of over 2,000 LLM-based synthetic personas. We compare these forecasts both to realized economic outcomes and, perhaps more interestingly, to the forecasting patterns of a panel of human experts. Third, we conduct a controlled ablation experiment to isolate the specific contribution of persona descriptions.
We observe two main results: while LLMs can achieve competitive forecasting performance alongside human experts, the sophisticated persona descriptions contribute negligible improvements over simple baseline prompts. This result has significant implications for practitioners, suggesting that computational resources may be better allocated to ensemble methods or model improvements rather than elaborate persona engineering. 
These insights contribute to the growing understanding of LLMs as \enquote{synthetic forecasters} while providing practical guidance for central banks and financial institutions considering AI-augmented forecasting systems. Our results suggest that effective LLM-based forecasting may depend more on robust data integration and model architecture than on prompt engineering.
%Most importantly, our out-of-sample evaluation on 2024-2025 economic events---completely unseen during model training--- demonstrates that these models can extrapolate beyond their training data to maintain competitive accuracy on genuinely novel economic developments.

The remainder of this paper is organized as follows. Section \ref{sec:related} reviews related work on LLM forecasting and prompt engineering. Section \ref{method} describes our data and experimental setup, including the ECB Survey of Professional Forecasters, and the persona dataset. Section \ref{sec:metrics} outlines our evaluation methodology. Section \ref{sec:results} presents results on the effectiveness of persona prompting, forecasting accuracy and human vs AI panel performance. Section \ref{sec:discussion} discusses future work, and Section \ref{sec:conclusion} concludes.

\begin{figure}[h!]
\includegraphics[width=0.488\textwidth]{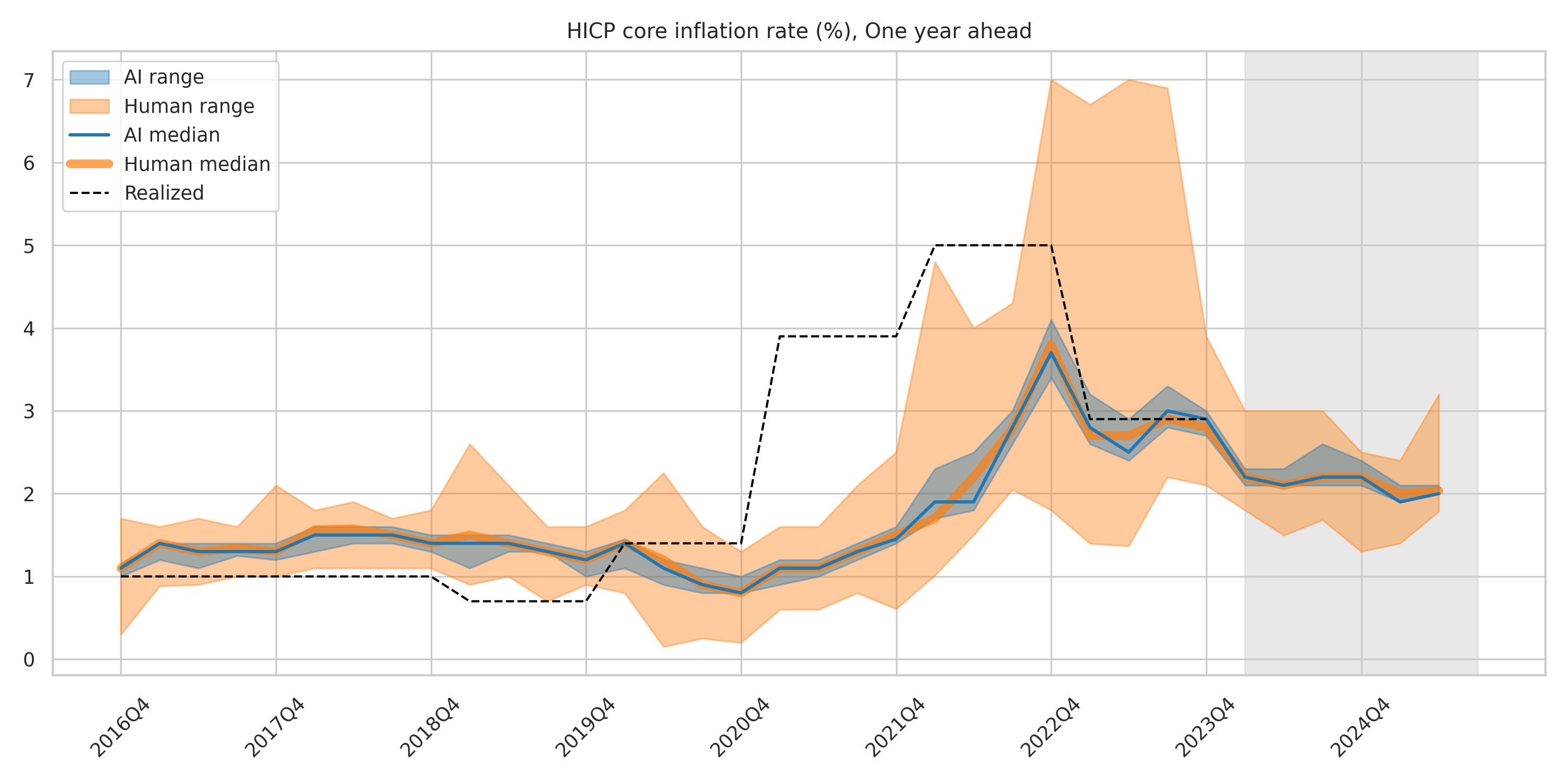}
\Description[Forecast comparison for euro area core inflation]{Line chart showing quarterly euro area core inflation forecasts from 2016Q4 to 2025Q2. 
It includes three sets of data: realized values (black dashed line), human expert median forecasts with a shaded distribution range (orange), and AI-generated median forecasts with a narrower distribution range (blue). Both AI and human  forecasts follow the realized trend closely. The AI forecast range is visibly narrower than  the human range, especially during periods of high inflation volatility. Out-of-sample quarters are shaded in gray.}
\caption{AI and human forecasters achieve remarkably similar accuracy across key macroeconomic variables. Time series comparison of realized outcomes (black), human expert forecasts from the ECB Survey of Professional Forecasters (orange), and AI forecasts using 2,368 synthetic personas (blue) for euro area core inflation (2016-2025). Despite using a variety of persona descriptions, LLM predictions converge to a much narrower forecast distribution compared to the human experts.}
\label{range-example}
\end{figure}

\section{Related work}\label{sec:related}
% \subsection{Forecasting with Large Language Models}
% \subsection{Persona prompting and prompt engineering}
\paragraph{Macroeconomic forecasting with Large Language Models}
The application of large language models to macroeconomic forecasting has emerged as a significant research area at the intersection of artificial intelligence and economics. Recent studies have explored direct applications of LLMs to a variety of forecasting tasks. Several studies share our focus on macroeconomic variables: \citet{carriero_macroeconomic_2024} examined LLM performance on macroeconomic time series, \citet{bybee_surveying_2023} fed Wall Street Journal articles to an LLM to predict financial and macroeconomic variables, while \citet{faria-e-castro_artificial_2024} demonstrated that Google's PaLM model could generate competitive inflation forecasts. 
%In financial market settings, Chen et al. (2022) applied BERT, RoBERTa, and OPT models to Thomson Reuters news feeds for firm-level return prediction, and Kim et al. (2024) utilized ChatGPT-4 with financial statements for earnings forecasting.
Our work extends these existing studies primarily through our rigorous focus on persona prompting and the deployment of a panel of more than 2,000 ''synthetic forecasters'', which enables systematic comparison against both expert human panels and realized macroeconomic outcomes.
Beyond evaluating LLM performance in specific forecasting applications, ongoing research has investigated general methodological approaches to LLM-based forecasting. \citet{lopez-lira_memorization_2025} investigated memorization effects in LLM-based economic forecasts, while \citet{tan2024language} compared language models to traditional time series methods. This methodological investigation connects to parallel efforts developing rigorous frameworks for LLM usage in economics research, which has become increasingly important as the field matures \citep{llm-econometric, korinek_language_2023}.
\paragraph{Simulating surveys responses}
A related strand of research examines LLMs' capacity to simulate human survey responses and expert judgment, similar to our comparison of human and AI panels in the Survey of Professional Forecasters. \citet{zarifhonarvar_evidence_2024} investigated inflation expectations formation using generative AI, finding that LLMs replicate key behavioral patterns including the tendency to predict higher inflation than realized rates. \citet{horton_large_2023} explored LLMs as ''simulated economic agents'', while \citet{argyle2023out} demonstrated that language models can replicate human samples in political surveys. \citet{geng2024large} and \citet{fell2024energy} examined LLMs' ability to simulate social survey responses, though \citet{dominguez-olmedo_questioning_2024} identified important limitations regarding ordering and labeling biases in LLM survey responses.
Most directly relevant to our investigation, \citet{hansen_simulating_2025} conducted the first systematic replication of the U.S. Survey of Professional Forecasters using LLMs, demonstrating comparable accuracy between AI and human forecasts. Their approach employed manually crafted forecaster personas based on SPF participant characteristics, in contrast with our larger extraction of persona prompts from the PersonaHub dataset.
\paragraph{Persona prompting}
A parallel line of research has examined the effects of prompt design and expert personas on LLM behavior. LLMs are highly sensitive to prompt content, structure, and formatting, as demonstrated by \citet{sclarquantifying}, who show that even subtle variations in prompt phrasing can lead to large shifts in model output. This sensitivity complicates the interpretation of results in applied settings, where changes in tone, emphasis, or structure may unintentionally influence forecast quality. One strategy to improve LLM performance on reasoning tasks involves persona prompting or role-play. \citet{kong-etal-2024-better} introduces a role-play prompting method in which LLMs are instructed to assume the identity of domain experts. This approach improves zero-shot performance across multiple benchmarks and has inspired further exploration of expert simulation in applied domains, including economics.
Persona prompting has also been used in macroeconomic forecasting experiments \citep{hansen_simulating_2025, zarifhonarvar_evidence_2024}. \citet{zarifhonarvar_evidence_2024} prompts LLM to adopt distinct persona attributes, such as political orientation, background, and socioeconomic characteristics. This induces realistic behaviors like partisan bias in inflation expectations, which mirror behaviors observed in actual human surveys. \citet{hansen_simulating_2025} construct detailed forecaster personas by manually gathering background data on SPF participants, including education, institutional affiliations, professional roles, and degrees. The study finds that removing these personas, i.e. replacing them with a generic forecaster prompt, leads to measurable drops in forecast accuracy, highlighting the value of role-based prompting. Interestingly, this is partially in contrast with our findings as presented in Section \ref{sec:results}.

\begin{table*}[htb]
\centering
\caption{Examples of economics-related blurbs contained in the PersonaHub dataset, evaluated on the three dimensions relevant to our study: EU-centrality, neutrality and monetary policy depth. Only one meets all three criteria and was retained for the experiments.}
\captionsetup{justification=centering}
\renewcommand{\arraystretch}{1.1}

\begin{tabular}{
    >{\raggedright\arraybackslash}m{0.6\linewidth}  % Blurb: vertical + left
    >{\centering\arraybackslash}m{1.6cm}             % EU centrality: centered both
    >{\centering\arraybackslash}m{1.5cm}             % Neutrality: centered both
    >{\centering\arraybackslash}m{2cm}               % MP depth: centered both
}
\hline
\textbf{Persona blurb (truncated)} & \textbf{EU centrality} & \textbf{Neutrality} & \makecell{\textbf{Expertise}} \\
\hline
\small
\textit{``A financial economist who specializes in the analysis of economic cycles and monetary policy. This person is interested in the degree of synchronisation of the euro area's economic cycle with that of the US, and how this affects the implementation of monetary policies. They are also interested in the factors that contribute to the degree of synchronisation and how they differ between the euro area and the US.''}
&\scalebox{1.5}{$\checkmark$} & \scalebox{1.5}{$\checkmark$} & \scalebox{1.5}{$\checkmark$} \\
\small
\textit{``A global economist with Bank of America Merrill Lynch, with expertise in inflation and deflation, particularly in the context of the US and Europe. They are optimistic about the potential for economic growth in the US, but also recognize the potential for shocks that could trigger deflation.''}
& \scalebox{1.5}{\ding{55}} & \scalebox{1.5}{\ding{55}} & \scalebox{1.5}{$\checkmark$} \\
\small
\textit{``A technologist who is skeptical of the effectiveness of information technology in stimulating economic growth. This persona believes that technology must be implemented and funded in order to generate economic growth. They also believe that the central bank's role in manipulating financial markets is a major impediment to economic growth.''}
& \scalebox{1.5}{\ding{55}} & \scalebox{1.5}{\ding{55}} & \scalebox{1.5}{\ding{55}}\\
\hline
\end{tabular}
\label{tab:persona_examples}

\end{table*}
\vspace{1em}
\section{Data and experimental setup}\label{method}
\subsection{Persona dataset}
Starting from the Persona Hub corpus \cite{ge_scaling_2024}, which is publicly available and contains $\approx$ 370 million descriptions of domain experts (\(p_i\)), we implement a multi-stage filtering pipeline to extract only the items relevant to our study. The \textit{Persona Hub} dataset includes highly heterogeneous personas—from lawyers to artists and policy analysts—thus requiring several layers of domain-specific filtering. The steps below were applied in the order presented:

% dire che questi step di filtering sono stati eseguiti in ordine
% criterio 1: per escludere le personas del tutto estranee --> dire che il persona hub contiene personas variegate --> vogliamo personas che abbiano il domain knowlegde giusto una prima esclusione massiccia 
% DA FARE: aggiungere in appendice anche 3 esempi di personas totalmente irrilevanti
% criterio 2: abbastanza self-explanatory?!
% criterio 3: varietà di personas --> potrebbero esserci descrizioni quasi identiche, non ci giova tenerle tutte
% criterio 4: abbiamo preso un sample e annotato manualmente (2 annotatori) e controllato che quello che gpt stesse facendo fosse corretto/sensato (avevamo calcolato una metrica ?!) --> perché abbiamo scelto questi 3 criteri? 1) con focus su BCE visto che si tratta di un survey della BCE 2) no bias, 3) ... 
\begin{enumerate}
    \item \textbf{Keyword search and domain filtering}: retain entries containing $\geq 2$ tokens from a lexicon of terms related to monetary policy and the ECB (e.g., \enquote{central banking}, \enquote{monetary policy}, \enquote{Governing Council}). The dataset also includes four \textit{domain} columns that were used as additional filtering dimensions. This step broadly excludes irrelevant personas while preserving those in macroeconomic or financial domains.
    \item \textbf{Name filtering}: using a named entity recognition (NER) algorithm, remove any description that directly mentions individuals (e.g., \enquote{Mario Draghi}) to avoid role-play blurbs. After the first two steps, the dataset was reduced to $\approx 200,000$ personas.
    \item \textbf{Duplicate removal}: drop overly similar personas by computing vector embeddings\footnote{\texttt{sentence-transformers/all-mpnet-base-v2}.} and discard those with cosine similarity scores \(\geq 0.90\), ensuring a diverse and representative subset of domain experts. This step reduced the dataset to $\approx 43,000$ personas.
    \item \textbf{Zero-shot relevance rating}: prompt an LLM (\texttt{gpt-4o\allowbreak-mini}) to evaluate each persona based on three independent binary criteria: \textit{EU-centrality}, \textit{neutrality}, and \textit{monetary policy depth} (see Appendix A for the full prompt). Keep only those personas that satisfy all three (i.e., $\textit{score}_{p_i} =3$). For increased robustness, we run the evaluation three times with temperature \texttt{T = 1} and apply majority voting to determine the final decision. To validate the model's reliability, we randomly sampled 50 personas and had two human annotators rate them manually. Cohen’s kappa scores between human ratings and GPT ratings ranged from $0.61$ to $0.81$ across the three criteria, indicating substantial agreement and confirming that the model selections are consistent with human judgment.

    % mostrare il prompt usato (da mettere in appendice)
    % dire che abbiamo selezionato randomicamente un sample di 50 persona descriptions e annotato manualmente (2 annotatori). per ogni criterio abbiamo calcolato il cohen's kappa score (mettere un appunto in footnote) e ritenuto sufficientemente buono il modello nel fare questa selezione perché lo score era in 0.61 e 0.81 per tutti e 3 i criteri --> substantial agreement. That means your raters (or model and true labels) agree much more often than chance would predict.
    % kappa score: it tells you how much two raters (or models and ground truth) agree beyond what you’d expect by chance
    
\end{enumerate}
Given the size of the starting dataset, a multi-step pipeline is necessary to exclude candidate prompts that would pass the initial keyword-based screening but are ultimately unsuitable for our study. This strategy yields a candidate pool \(P^{\star}\) containing \textbf{2368} biographies, representing a highly selective filter that retains approximately six personas per million from the initial dataset. Appendix B provides examples of economics-related persona descriptions from the PersonaHub dataset, illustrating how personas were evaluated and selected according to EU-centrality, neutrality, and monetary policy expertise.
%Appendix \ref{app:personas} further details our filtering process by showing how blurbs are evaluated and selected based on EU-centrality, neutrality, and monetary policy depth.

%to illustrate this point: (i) one ideal description retained after all screening stages, 
%(ii) one eliminated for its US focus despite hitting monetary policy keywords, and (iii) one excluded for being too topic-specific.

\subsection{ECB Survey of Professional Forecasters} 
The European Central Bank's \textit{ Survey of Professional Forecasters} (SPF) is a quarterly survey that collects forecasts from a panel of experts on key euro area macroeconomic indicators. These include HICP (Harmonised Index of Consumer Prices) inflation, real GDP growth, and the unemployment rate.
The survey covers multiple forecasting horizons: the current calendar year, the next year, two years ahead, rolling horizons, and a five-year outlook. Respondents provide both point forecasts and probability distributions.
For this study, we consider a total of 50 SPF rounds from 2013Q1 to 2025Q2 and only focus on four of the macroeconomic variables tracked by the survey (see Table \ref{tab:data_overview} for an overview of the dataset).

\begin{table*}[t]
\centering
\caption{Intra-panel dispersion comparison between AI persona-based and human forecasters. Values represent the median dispersion, measured with standard deviation (SD) and inter-quartile range (IQR), across all survey rounds. AI forecasts consistently exhibit substantially lower dispersion than human forecasters across all variables and horizons, while human forecasters display broader disagreement patterns typical of professional survey panels.}
\label{tab:variance_comparison}

\begin{subtable}[t]{0.48\textwidth}
\centering
\caption{HICP and HICPX}
\begin{tabular}{llcccc}
\toprule
Variable & Horizon & \multicolumn{2}{c}{SD} & \multicolumn{2}{c}{IQR} \\
\cmidrule(lr){3-4} \cmidrule(lr){5-6}
        &         & AI & Human & AI & Human \\
\midrule
HICP   & CY   & 0.030 & 0.164 & 0.000 & 0.200 \\
       & CY+1 & 0.041 & 0.243 & 0.000 & 0.300 \\
       & CY+2 & 0.039 & 0.255 & 0.000 & 0.270 \\
       & LT   & 0.009 & 0.210 & 0.000 & 0.215 \\
HICPX  & CY   & 0.045 & 0.170 & 0.006 & 0.200 \\
       & CY+1 & 0.045 & 0.242 & 0.000 & 0.275 \\
       & CY+2 & 0.040 & 0.253 & 0.025 & 0.300 \\
       & LT   & 0.012 & 0.257 & 0.000 & 0.285 \\
\bottomrule
\end{tabular}
\end{subtable}
\hfill
\begin{subtable}[t]{0.48\textwidth}
\centering
\caption{rGDP and UNR}
\begin{tabular}{llcccc}
\toprule
Variable & Horizon & \multicolumn{2}{c}{SD} & \multicolumn{2}{c}{IQR} \\
\cmidrule(lr){3-4} \cmidrule(lr){5-6}
        &         & AI & Human & AI & Human \\
\midrule
rGDP   & CY   & 0.040 & 0.183 & 0.000 & 0.200 \\
       & CY+1 & 0.048 & 0.257 & 0.000 & 0.300 \\
       & CY+2 & 0.042 & 0.262 & 0.000 & 0.300 \\
       & LT   & 0.018 & 0.300 & 0.000 & 0.331 \\
UNR    & CY   & 0.047 & 0.194 & 0.050 & 0.177 \\
       & CY+1 & 0.046 & 0.296 & 0.000 & 0.300 \\
       & CY+2 & 0.043 & 0.417 & 0.000 & 0.419 \\
       & LT   & 0.024 & 0.582 & 0.000 & 0.700 \\
\bottomrule
\end{tabular}
\end{subtable}
\end{table*}

\paragraph{Training data leakage}
When trying to simulate responses to past events, such as past inflation, it is essential to consider potential data leakage issues which could lead to the ''memorization problem'' \citep{lopez-lira_memorization_2025, llm-econometric}. We build an out-of-sample training set including SPF rounds from 2024-Q1 to 2025-Q2. This time period contains six editions of the ECB Survey of Professional Forecasters. These surveys occur after the training cutoff date for the \textsc{gpt-4o} model\footnote{Specifically, we prompt \textbf{gpt-4o-2024-11-20}.}, which is the object of this study \cite{openai2024gpt4o}, ensuring the absence of data leakage. 
\subsection{Prompt architecture}\label{prompt}
The central element of each LLM prompt in this study is the persona description (\(p_i\)), which we evaluate for its impact on task performance. In addition to the persona, each prompt includes a set of standardized components necessary to perform the two experimental tasks. For each experimental call, the LLM receives:  
\begin{itemize}
    \item \textbf{System rules}: date anchoring ($T_s$), instructions for output formatting.
    \item \textbf{Persona blurb} \(p_i\in P^{\star}\), unedited except for the controlled experiment settings---see below.
    \item \textbf{Monetary policy context}: the full text of the latest ECB press release at time $T_s$, plus a macro snapshot\\
    \((\pi^{\text{HICP}},\;gdp^{\text{now}},\;\dots)\).
    \item \textbf{Task instruction}: 

   \textit{“Provide your forecasts for euro area HICP inflation, real GDP growth, and unemployment rate for the current quarter (t) the following time horizons t+1, t+2, t+3, t+4 (...) Format your response as numerical values: 'HICP (t): X.X\%, HICP (t+1): X.X\%, ...' etc.”}.

\end{itemize}
To isolate the effect of the persona selection on performance, we test two variations of the persona blurb: (i) \textit{raw text}: unedited, full \(p_i\); (ii) \textit{empty persona}: the persona description block is omitted, providing only the monetary policy context and the task instructions.
An example of the full prompt can be found in Appendix C.
\begin{figure*}[h]
    \centering
    
    % First row
    \begin{subfigure}[b]{0.47\textwidth}
        \centering
        \includegraphics[width=\textwidth]{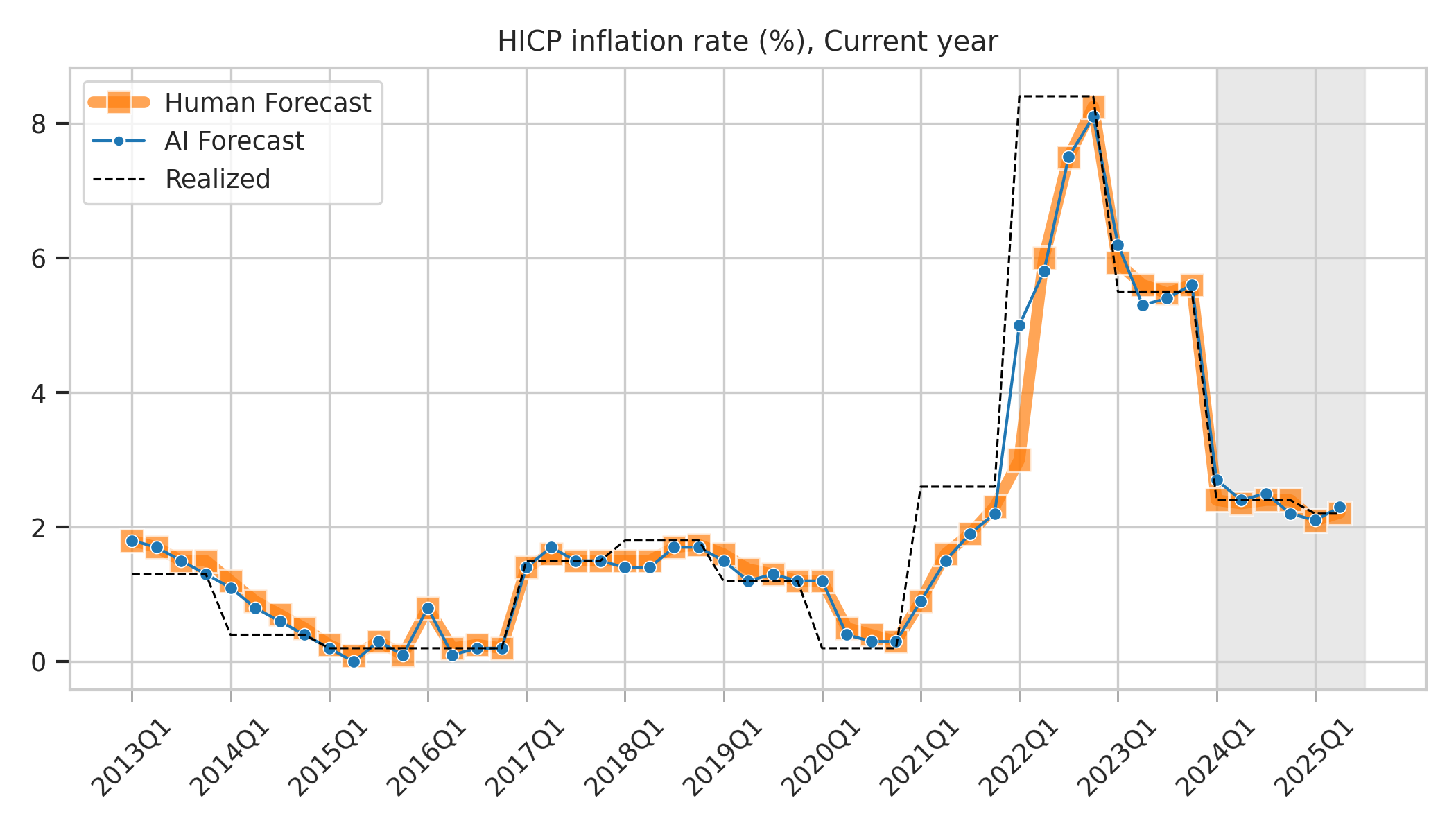}
      %  \caption{Caption for subplot (a)}
        \label{fig:sub1}
    \end{subfigure}
    \hfill
    \begin{subfigure}[b]{0.47\textwidth}
        \centering
        \includegraphics[width=\textwidth]{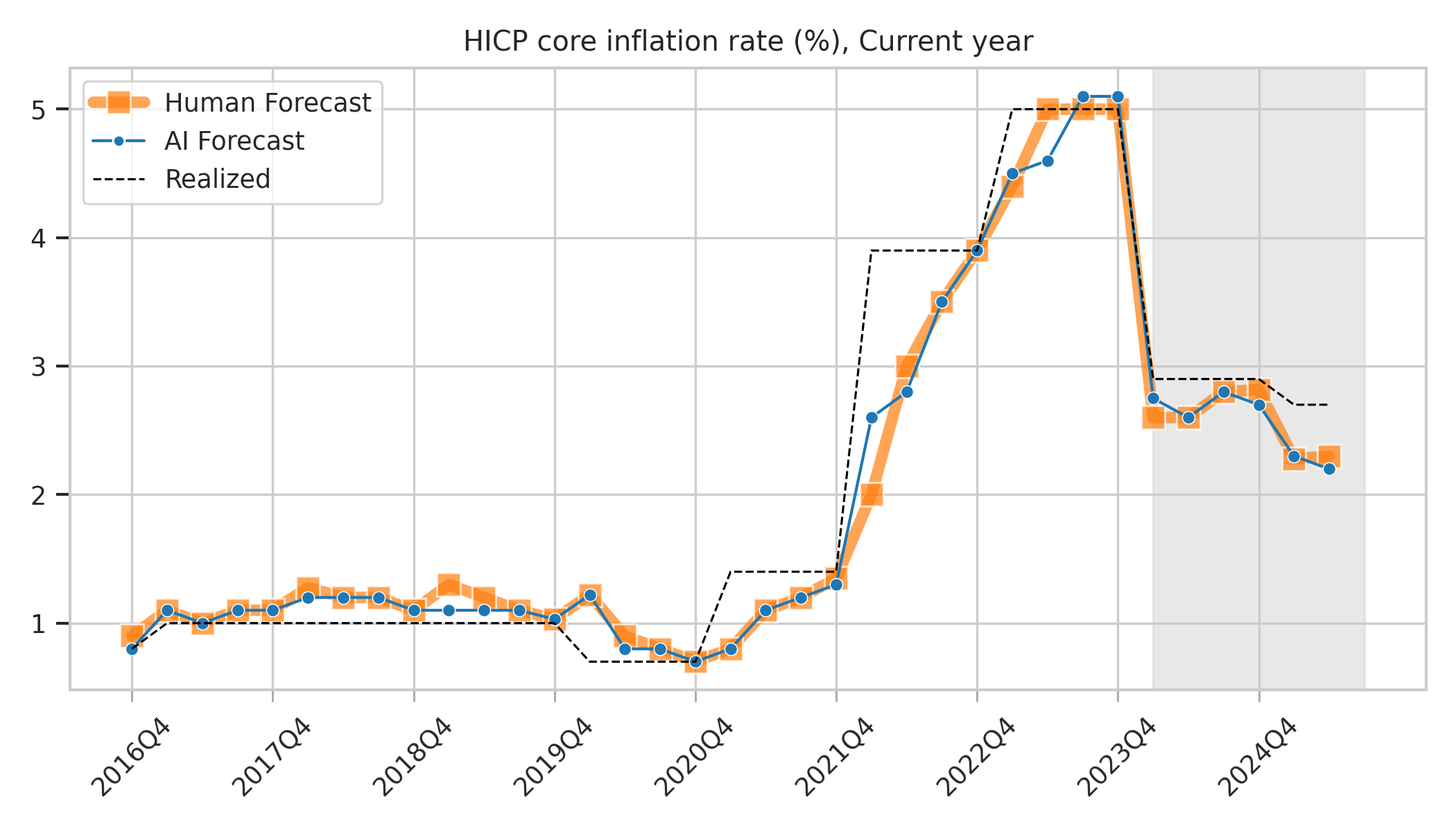}
   %     \caption{Caption for subplot (b)}
        \label{fig:sub2}
    \end{subfigure}
    \vspace{-0.7em}
    
    % Second row
    \begin{subfigure}[b]{0.47\textwidth}
        \centering
        \includegraphics[width=\textwidth]{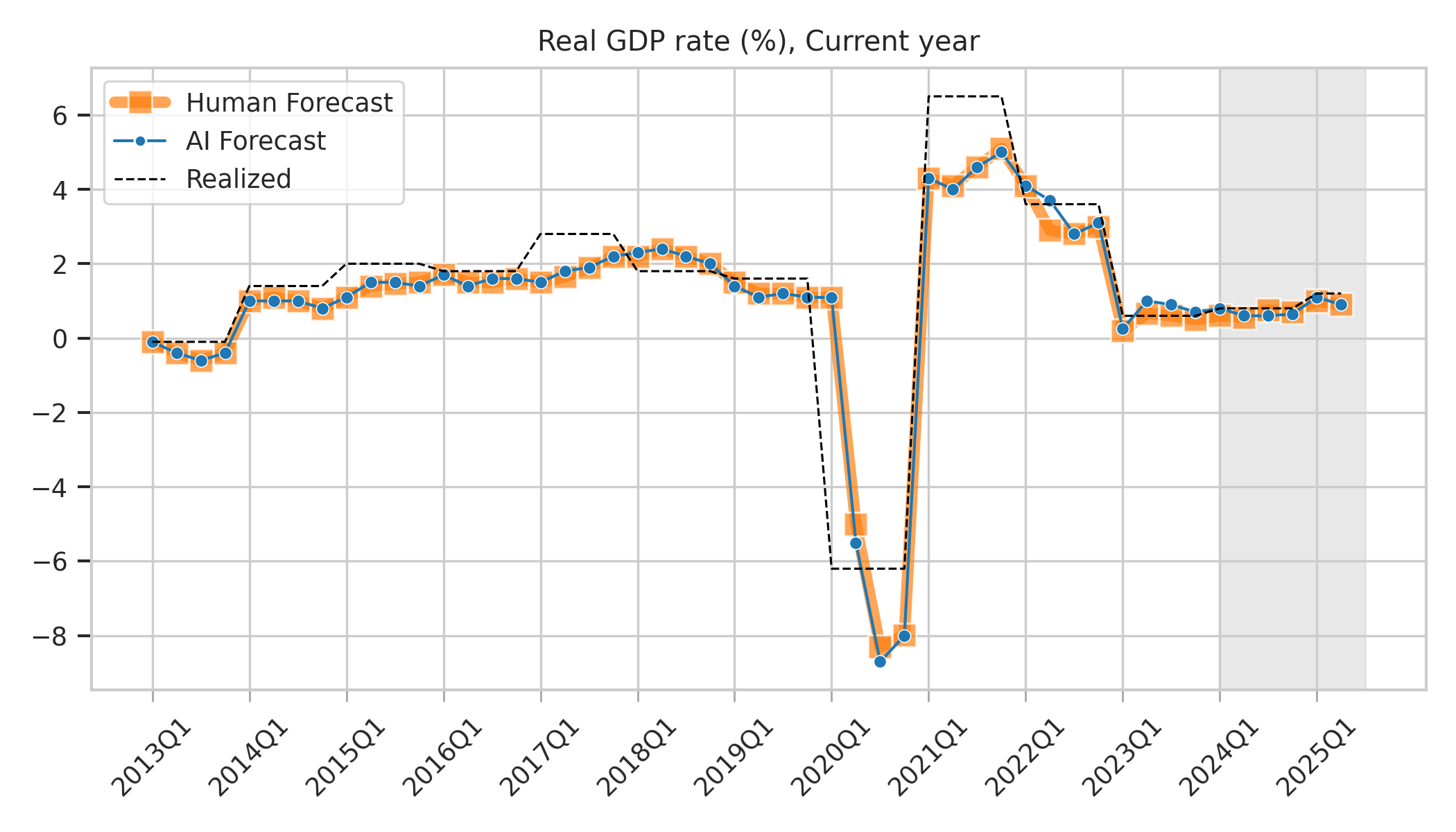}
      %  \caption{Caption for subplot (c)}
        \label{fig:sub3}
    \end{subfigure}
    \hfill
    \begin{subfigure}[b]{0.47\textwidth}
        \centering
        \includegraphics[width=\textwidth]{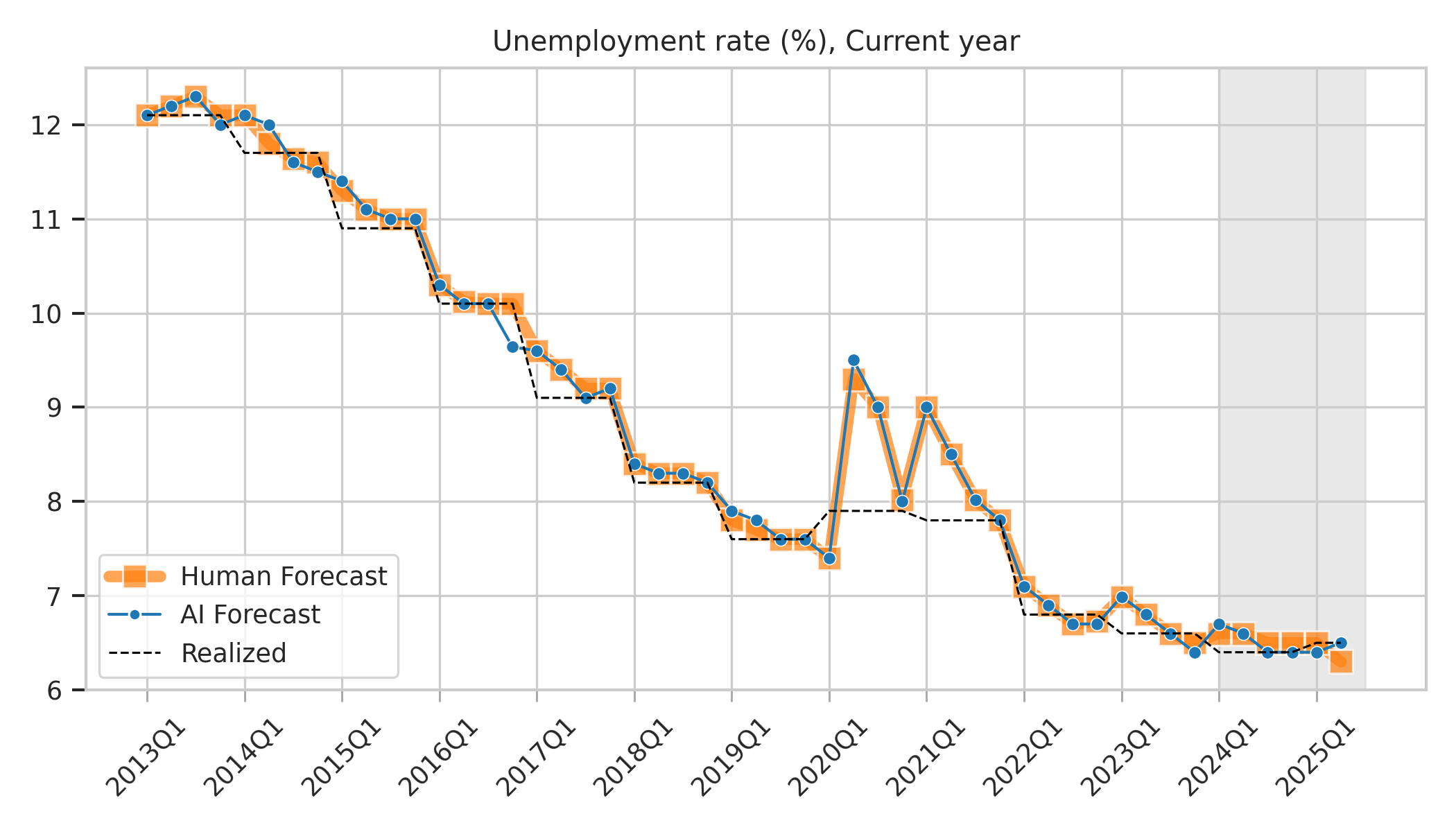}
  %      \caption{Caption for subplot (d)}
        \label{fig:sub4}
    \end{subfigure}
    \vspace{-1.2em}
    
    \Description[AI vs human forecasts for key macro variables] {Four-panel figure with four line charts comparing AI and human forecasts for euro area macroeconomic indicators from 2013Q1 to 2025Q2. Each subplot represents a different variable: (a) HICP inflation rate, (b) HICP core inflation rate, (c) real GDP growth, and (d) unemployment rate. For each variable, the chart displays the realized values as a black dashed line, the AI persona-based median forecast as a blue line, and the human forecast median from the ECB Survey of Professional Forecasters as an orange line. All variables are shown for the current-year forecast horizon. The gray-shaded area indicates the out-of-sample evaluation period, 
    beginning in 2024Q1. Across all variables, AI and human forecasts track realized outcomes closely, with visible differences during 
    volatile periods.}
    \caption{Comparison of AI persona-based and human forecasts for current-year horizon across four ECB-SPF variables (2013-2025): (a) HICP inflation, (b) HICP core inflation, (c) Real GDP growth, and (d) Unemployment rate. Gray shaded regions indicate out-of-sample evaluation period. AI-generated median forecasts often, but not always, match human forecasts; this occurs both in the in-sample and out-of-sample surveys.}
    \label{fig:main}
\end{figure*}
\subsection{Scoring metrics}\label{sec:metrics}

%\begin{multline}
% v \in \{\text{HICP, HICPX,} \\
% \text{rGDP, UNR}\}
% \end{multline}
% Evaluation is carried out \emph{match-by-match}, where a \emph{match} is the unique
% combination of survey round $r$, macro variable\\
% $v\in\{\text{HICP, HICPX, rGDP, UNR}\}$, and forecast horizon
% $h\in\{\text{t0,t1,t2,lt}\}$.
Evaluation is carried out \emph{match-by-match}, where a \emph{match} is the unique
combination of survey round $r$, macro variable
$v \in \{\text{HICP,}\allowbreak\ \text{HICPX,}\allowbreak\ \text{rGDP,}\allowbreak\ \text{UNR}\}$, and forecast horizon
$h \in \{\text{t0, t1, t2, lt}\}$.
For each match we collapse the $\sim$2\,000 persona completions of a given
LLM to a single forecast---its cross-sectional \emph{median}---and compare it
with the published SPF-panel median.  With the first real-time annual average
that becomes available after the reference year closes, we form the absolute errors
\[
e^{\text{AI}}_{rvh}=|\hat{y}^{\text{AI}}_{rvh}-y_{rvh}|,\qquad
e^{\text{H}}_{rvh}=|\hat{y}^{\text{SPF}}_{rvh}-y_{rvh}|.
\]

\paragraph{Point accuracy.} For each variable and horizon, we measure forecast accuracy against the realized yearly data with the mean-absolute error (MAE)
\[
\text{MAE}^{\text{panel}}_{vh}
   =\frac{1}{n_{vh}}\sum_{r=1}^{n_{vh}}e^{\text{panel}}_{rvh},
\]
with $n_{vh}$ being the number of available rounds. The score is reported in Table~\ref{tab:forecast_accuracy}. The lower of the two numbers is
bold-faced.

\paragraph{Panel disagreement.} For each round we measure cross-sectional dispersion with the
inter-quartile range \(\text{IQR}_{rvh}=q_{75}(\hat y_{\bullet rvh})-q_{25}(\hat y_{\bullet rvh})\), computed separately for personas and for human respondents. Table
\ref{tab:variance_comparison} reports the \emph{median} IQR and variance across rounds.
\paragraph{Relative performance (win–share).}
For every match, we record
\(
\mathrm{win}_{rvh}=\mathbf 1\!\{e^{\mathrm{AI}}_{rvh}<e^{\mathrm{H}}_{rvh}\},
\)
so that \(\mathrm{win}=1\) denotes an AI victory over the SPF median.
Let \(w_{vh}=\sum_{r}\mathrm{win}_{rvh}\) and
\(n_{vh}\) be the number of matches for variable \(v\) and horizon \(h\).
The \emph{win-share}
\[
\bar w_{vh}=w_{vh}/n_{vh}
\]
is reported alongside two p-values that address distinct questions:

\begin{itemize}[leftmargin=1.4em,itemsep=0pt]
\item \textbf{One-tailed ($H_{0}\!:\Pr(\mathrm{win})=0.5$ vs.\ $H_{A}\!:\Pr(\mathrm{win})>0.5$).}\\
      Answers “is the AI panel \emph{strictly better} than humans?”.
\item \textbf{Two-tailed} tests the symmetric alternative
      $H_{A}\!:\Pr(\mathrm{win})\neq0.5$ and answers
      “is there \emph{any} systematic difference in accuracy?”.
\end{itemize}

The same procedure is applied both to the \emph{panel-median} AI forecast and
to every \emph{individual persona}:

\begin{enumerate}[label=(\roman*)]
\item \textbf{In-sample rounds} (\(2013\text{Q}1\)–\(2023\text{Q}4\);
      up to \(n_{vh}=44\)).  
      We approximate the null distribution by Monte-Carlo:
      \(N=10{,}000\) artificial panels
      \(W_{j}^{*}\sim\mathrm{Binom}(n_{vh},0.5)\).
      \begin{align}
          p^{(1)}_{vh} &= N^{-1}\sum_{j}\mathbf{1}\{W^{*}_{j}\ge w_{vh}\}, \\
          p^{(2)}_{vh} &= 2\,\min\left(p^{(1)}_{vh}, N^{-1}\sum_{j}\mathbf{1}\{W^{*}_{j}\le w_{vh}\}\right).
      \end{align}
      With \(N=10{,}000\) the Monte-Carlo error never exceeds~0.005.
\item \textbf{Out-of-sample rounds} (\(2024\text{Q}1\)–\(2025\text{Q}2\);
      \(n_{vh}\le 6\)).  
      We use the exact binomial:
      \begin{align}
          p^{(1)}_{vh} &= \Pr\{W\ge w_{vh}\}, \\
          p^{(2)}_{vh} &= 2\,\min\{\Pr(W\ge w_{vh}),\Pr(W\le w_{vh})\},
      \end{align}
      where \(W\sim\mathrm{Binom}(n_{vh},0.5)\).
\end{enumerate}
Stars in every table refer to the \emph{one-tailed} p-value and mark
$^{*}\,p\le0.10$, $^{**}\,p\le0.05$, $^{***}\,p\le0.01$.
%Two-tailed p-values are provided in appendix

\begin{figure}[t]
\centering
\includegraphics[width=0.48\textwidth]{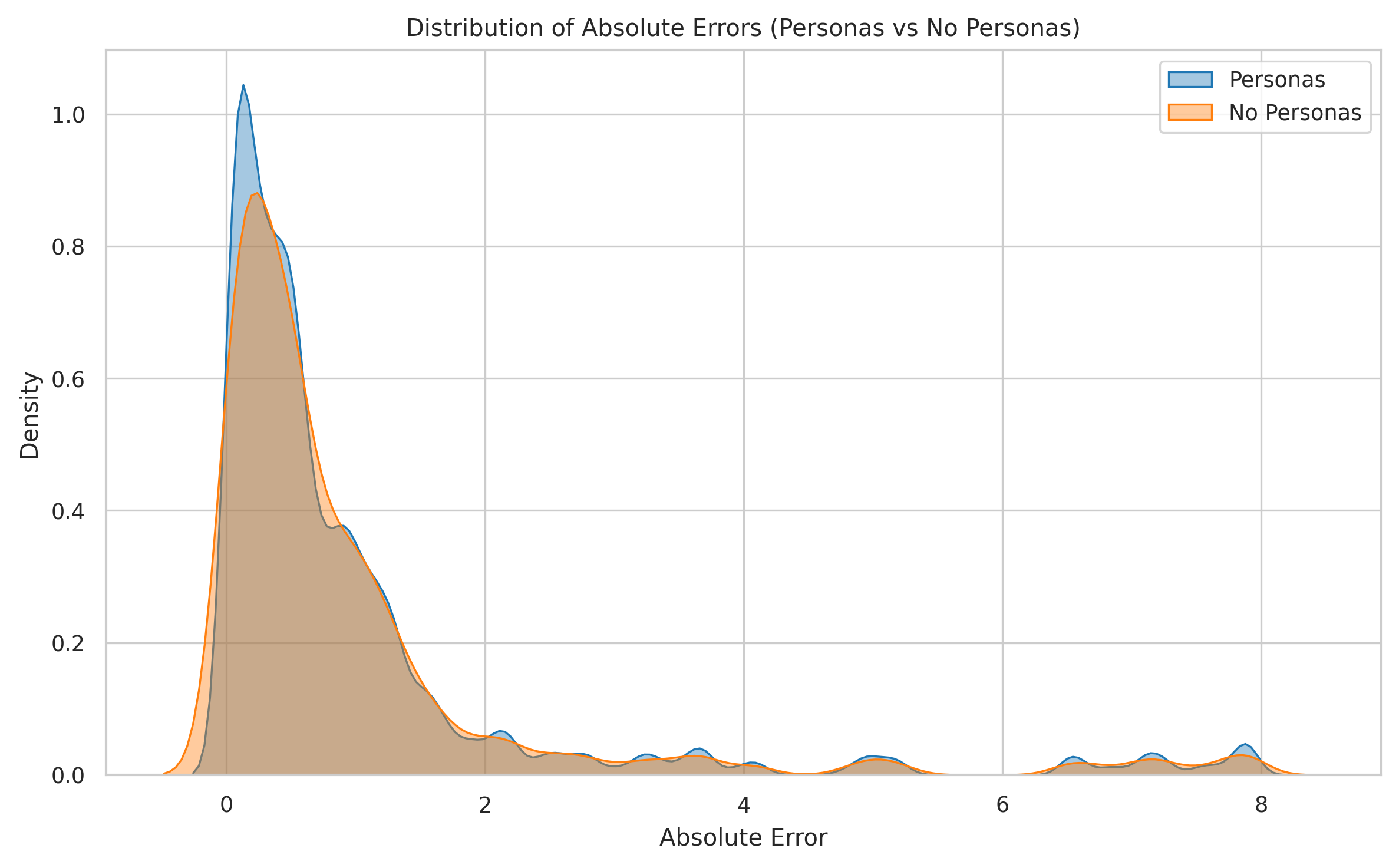}
\caption{Persona prompting yields statistically indistinguishable error distributions. Kernel density estimates of absolute forecast errors for GPT-4o with persona descriptions (blue) versus baseline prompts without personas (orange) across all variable-horizon-round combinations. The near-perfect overlap supports our null hypothesis ($t = -1.02$, $p = 0.31$; Kolmogorov-Smirnov $D = 0.05$, $p = 0.28$).}
\label{fig:persona_ablation}
\end{figure}
\begin{figure*}[!htbp]
\centering
\begin{subfigure}[b]{0.48\textwidth}
    \centering
    \includegraphics[width=\textwidth]{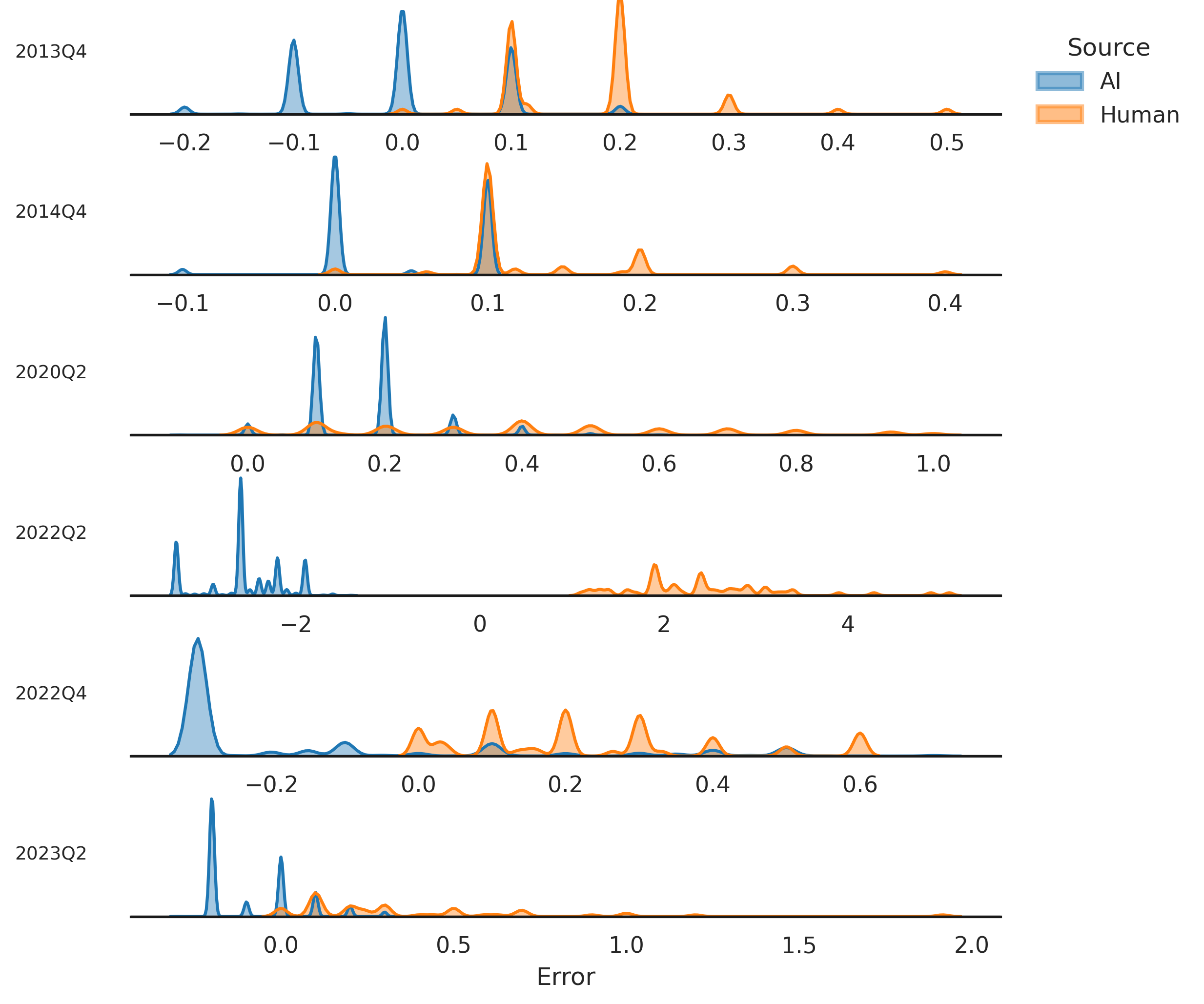}
    \caption{HICP inflation, Current year}
    \label{fig:ridge_current}
\end{subfigure}
\hfill
\begin{subfigure}[b]{0.48\textwidth}
    \centering
    \includegraphics[width=\textwidth]{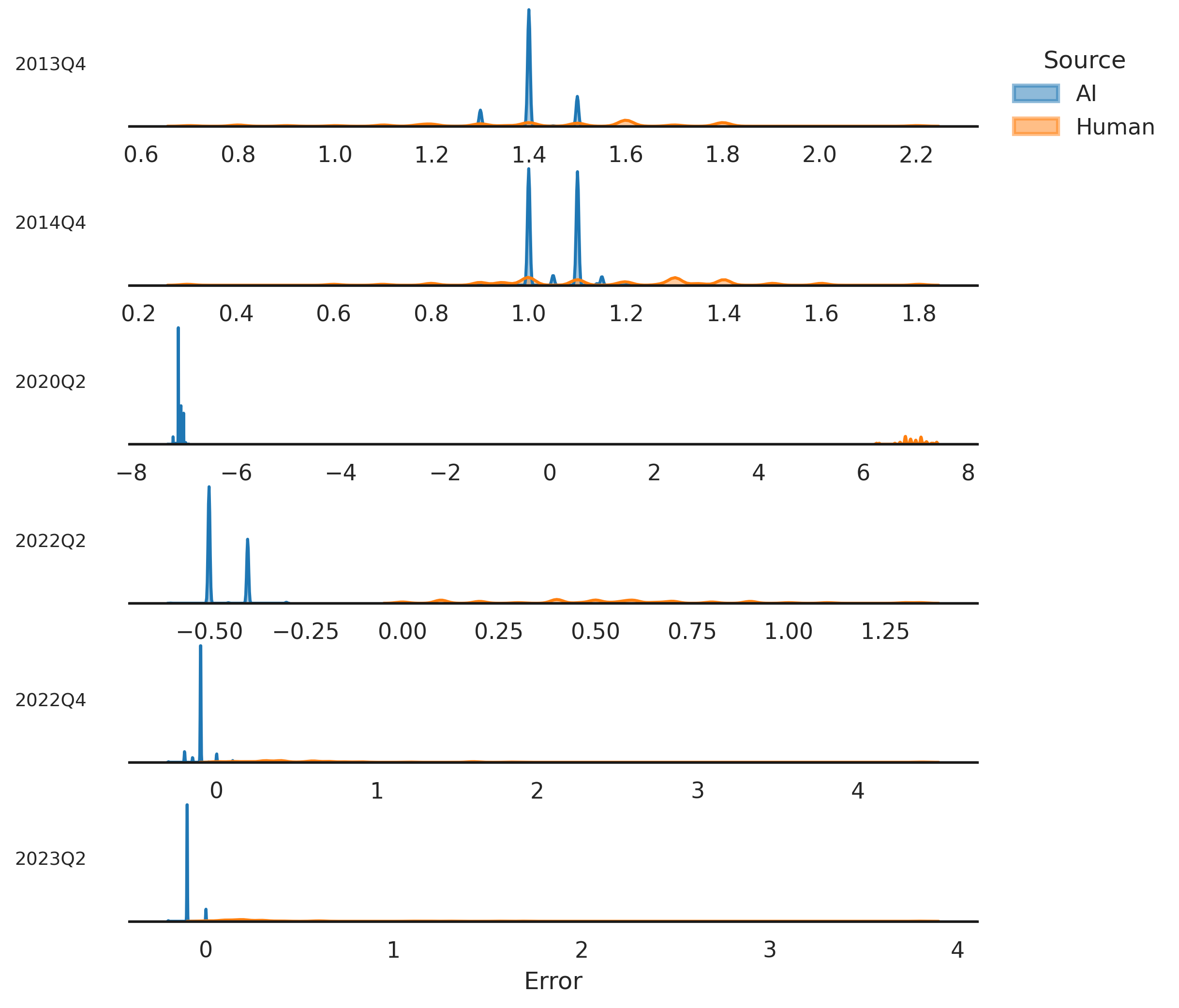}
    \caption{HICP inflation, Two years ahead}
    \label{fig:ridge_one_year}
\end{subfigure}

\vspace{0.5cm}

\begin{subfigure}[b]{0.48\textwidth}
    \centering
    \includegraphics[width=\textwidth]{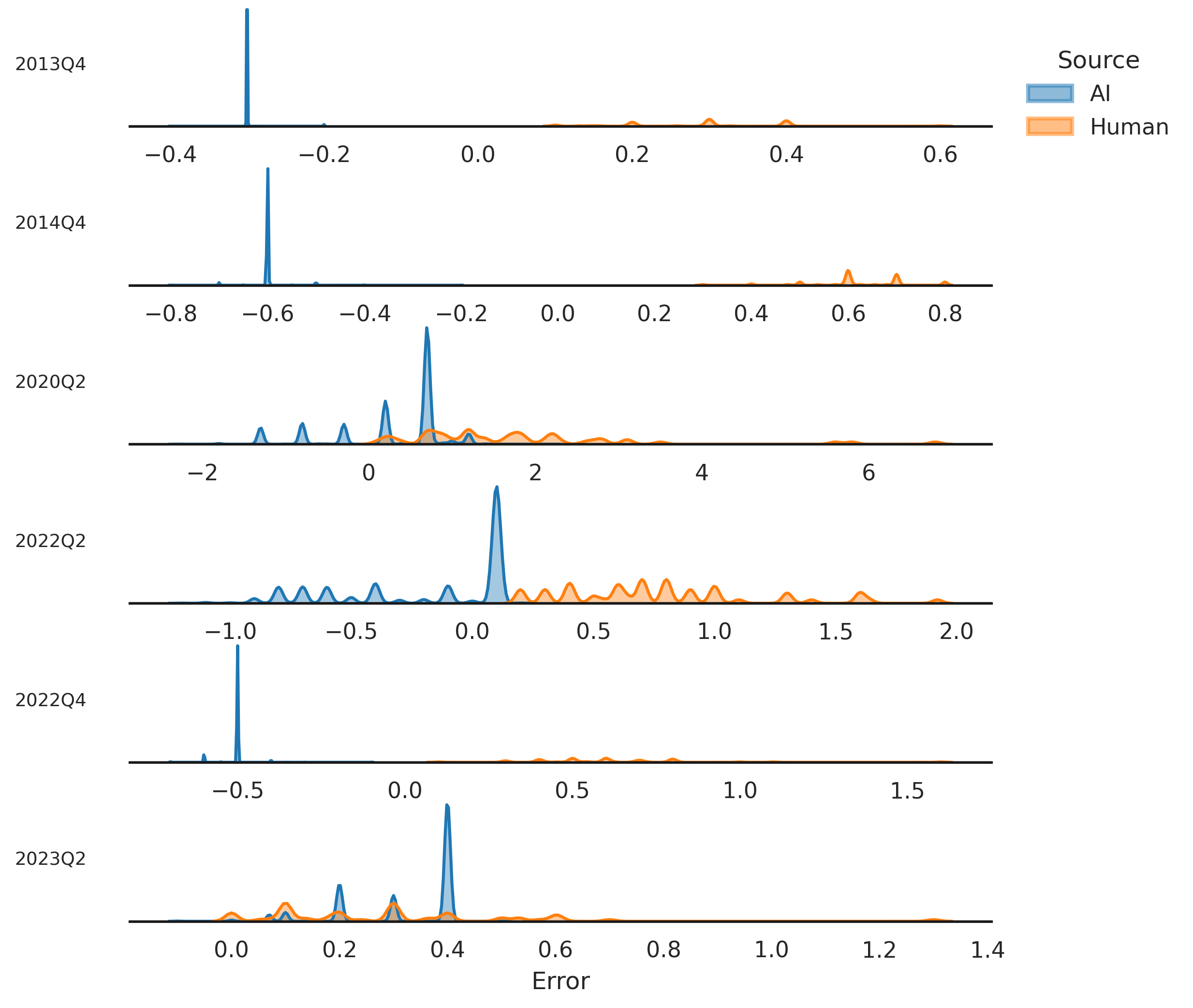}
    \caption{Real GDP growth, Current year}
    \label{fig:ridge_two_years}
\end{subfigure}
\hfill
\begin{subfigure}[b]{0.48\textwidth}
    \centering
    \includegraphics[width=\textwidth]{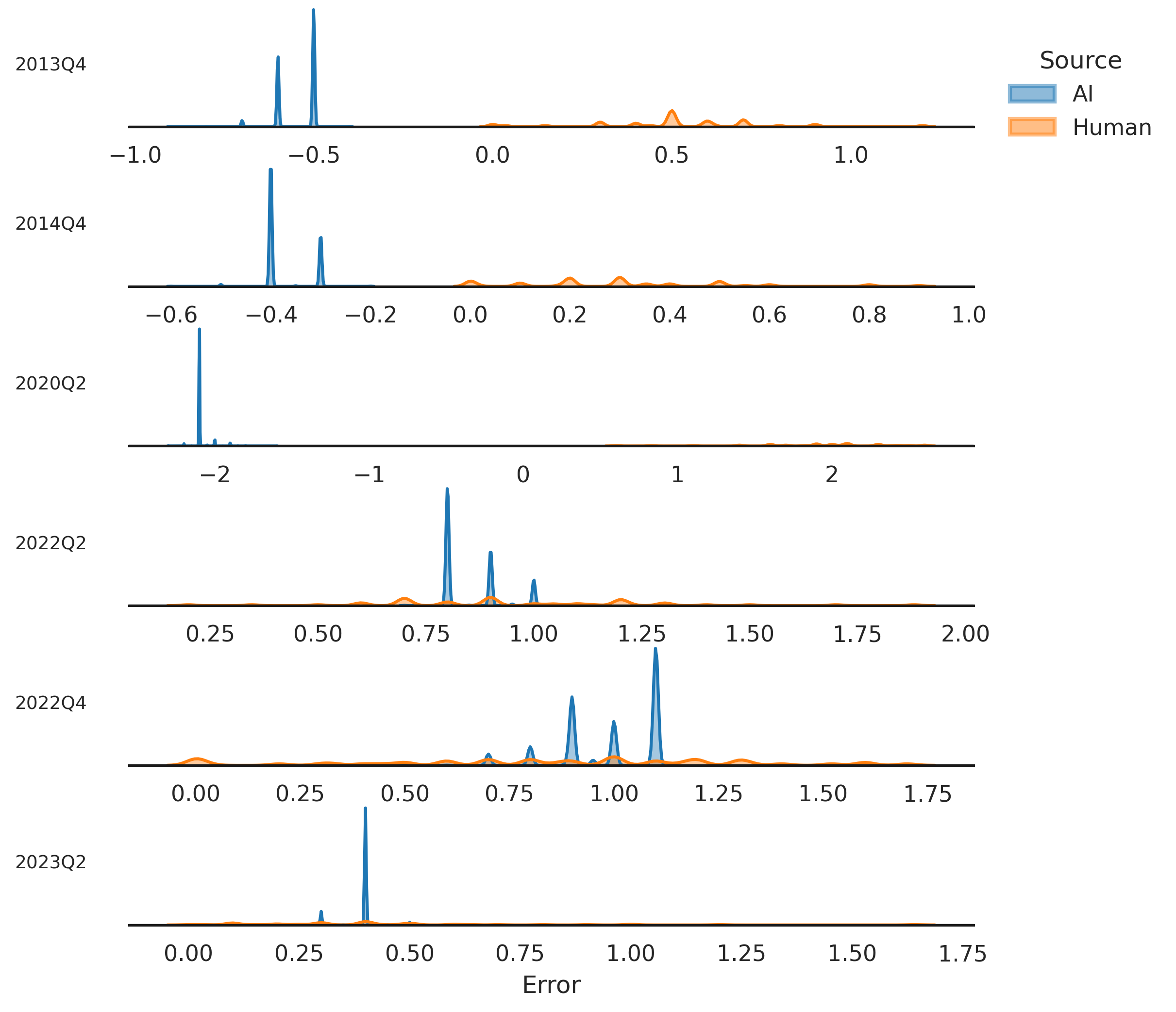}
    \caption{Real GDP growth, Two years ahead}
    \label{fig:ridge_long_term}
\end{subfigure}
\Description[Distribution of AI vs human forecast errors]{Four-panel figure showing kernel density plots of forecast errors for AI and human forecasts of euro area inflation. Each panel compares error distributions for AI forecasts (blue) and human ECB-SPF forecasts (orange). 
The top row displays errors for HICP inflation at two forecast horizons: current year (t0) and two years ahead (t2). 
The bottom row shows the same for real GDP growth (rGDP). In all panels, the AI forecast error distributions are more peaked and concentrated around zero, indicating lower dispersion compared to the broader and flatter human forecast errors.}
\caption{Distribution of forecast errors by variable and horizon. Each panel shows kernel density estimates of errors for AI forecasts (blue) and human SPF forecasts (orange) across selected survey rounds. Top row compares HICP inflation errors at current-year (t0) and two-years (t2) horizons. Bottom row shows the same comparison for real GDP growth. AI forecasts consistently exhibit lower dispersion and more concentrated error distributions than human forecasters across both inflation measures and forecast horizons.}
\label{fig:ridge_plots_inflation_rgdp}
\end{figure*}

\begin{table*}[t]
\centering
\caption{Forecast accuracy (MAE) comparing AI persona-based forecasts (median of 2,368 personas) versus human SPF medians across 50 ECB-SPF rounds (2013-2025). In-sample: 2013-2023, out-of-sample: 2024-2025. Bold indicates better performance. All errors in percentage points.}
\begin{tabular}{l*{12}{c}}
\toprule
& \multicolumn{8}{c}{In-sample} & \multicolumn{4}{c}{Out-of-sample} \\
\cmidrule(lr){2-9} \cmidrule(lr){10-13}
Horizon
& \multicolumn{2}{c}{CY} 
& \multicolumn{2}{c}{CY+1}
& \multicolumn{2}{c}{CY+2}
& \multicolumn{2}{c}{Long-term}
& \multicolumn{2}{c}{CY}
& \multicolumn{2}{c}{CY+1} \\
\cmidrule(lr){2-3} \cmidrule(lr){4-5} \cmidrule(lr){6-7} \cmidrule(lr){8-9} \cmidrule(lr){10-11} \cmidrule(lr){12-13}
 & AI & Human
 & AI & Human
 & AI & Human
 & AI & Human
 & AI & Human
 & AI & Human \\
\midrule
HICP & 0.20 & \textbf{0.19} & \textbf{0.95} & 1.00 & \textbf{1.02} & 1.03 & 0.75 & \textbf{0.74} & 0.10 & \textbf{0.01} & \textbf{0.13} & 0.19 \\
HICPX & \textbf{0.10} & \textbf{0.10} & \textbf{0.50} & \textbf{0.50} & \textbf{0.70} & 0.75 & \textbf{1.30} & \textbf{1.30} & \textbf{0.25} & 0.30 & \textbf{0.50} & \textbf{0.50} \\
rGDP & \textbf{0.50} & \textbf{0.50} & \textbf{0.50} & \textbf{0.50} & \textbf{0.60} & 0.90 & \textbf{0.85} & \textbf{0.85} & \textbf{0.17} & 0.20 & \textbf{0.20} & \textbf{0.20} \\
UNR & 0.20 & \textbf{0.10} & 0.47 & \textbf{0.42} & \textbf{0.70} & \textbf{0.70} & \textbf{1.00} & \textbf{1.00} & \textbf{0.05} & 0.15 & 0.05 & \textbf{0.02} \\
\bottomrule
\end{tabular}
\label{tab:forecast_accuracy}
\end{table*}
\begin{table*}[t]
\centering
\caption{Win share of SPF and AI-generated forecasts (\%). Win shares calculated as percentage of forecasting rounds where AI (or human) median strictly outperformed the other. Remaining percentage represents ties. $\ast\!\ast\!\ast$ $p \leq 0.01$}
\label{tab:win_share}
\begin{tabular}{lccrccrccrccrccr}
\toprule
Horizon
& \multicolumn{3}{c}{CY}
& \multicolumn{3}{c}{CY+1}
& \multicolumn{3}{c}{CY+2} 
& \multicolumn{3}{c}{Long-term} \\
\cmidrule(lr){2-4} \cmidrule(lr){5-7} \cmidrule(lr){8-10} \cmidrule(lr){11-13}
 & \shortstack{\rule{0pt}{1em}AI\\wins} 
 & \shortstack{\rule{0pt}{1em}Human\\wins} 
 & \shortstack{\rule{0pt}{1em}P-val}
 & \shortstack{\rule{0pt}{1em}AI\\wins} 
 & \shortstack{\rule{0pt}{1em}Human\\wins} 
 & \shortstack{\rule{0pt}{1em}P-val}
 & \shortstack{\rule{0pt}{1em}AI\\wins} 
 & \shortstack{\rule{0pt}{1em}Human\\wins} 
 & \shortstack{\rule{0pt}{1em}P-val}
 & \shortstack{\rule{0pt}{1em}AI\\wins} 
 & \shortstack{\rule{0pt}{1em}Human\\wins} 
 & \shortstack{\rule{0pt}{1em}P-val} \\
\midrule
\multicolumn{13}{l}{\textbf{In-sample}} \\
HICP  & 29 & \textbf{28} & $\ast\!\ast\!\ast$ & 30 & \textbf{31} & $\ast\!\ast\!\ast$ & \textbf{31} & 26 & $\ast\!\ast\!\ast$ & 5 & \textbf{11} & $\ast\!\ast\!\ast$ \\
HICPX & \textbf{31}  & 20  & $\ast\!\ast\!\ast$ & \textbf{30}  & 17  & $\ast\!\ast\!\ast$ & \textbf{26}  & 13  & $\ast\!\ast\!\ast$ & 1 & \textbf{5} & $\ast\!\ast\!\ast$ \\
rGDP  & 20  & \textbf{30} & $\ast\!\ast\!\ast$ & 16  & \textbf{34} & $\ast\!\ast\!\ast$ & 26 & \textbf{28} & $\ast\!\ast\!\ast$ & \textbf{17} & 9 & $\ast\!\ast\!\ast$ \\
UNR   & 13  & \textbf{32} & $\ast\!\ast\!\ast$ & \textbf{19}  & 17  & $\ast\!\ast\!\ast$ & \textbf{23} & 15  & $\ast\!\ast\!\ast$ & \textbf{14} & 9 & $\ast\!\ast\!\ast$ \\
\midrule
\multicolumn{13}{l}{\textbf{Out-of-sample}} \\
HICP  & 28 & \textbf{68} & $\ast\!\ast\!\ast$ & \textbf{70} & 25 & $\ast\!\ast\!\ast$ &  &  &  &  &  &  \\
HICPX & 32 & \textbf{40} & $\ast\!\ast\!\ast$ & \textbf{25} & 0  & $\ast\!\ast\!\ast$ &  &  &  &  &  &  \\
rGDP  & \textbf{53} & 25 & $\ast\!\ast\!\ast$ & 10 & \textbf{50} & $\ast\!\ast\!\ast$ &  &  &  &  &  &  \\
UNR & \textbf{47} & 37 & $\ast\!\ast\!\ast$ & 3 & \textbf{40}  & $\ast\!\ast\!\ast$ &  &  &  &  &  &  \\
\bottomrule
\end{tabular}
\end{table*}

\section{Results}\label{sec:results}
We report results obtained with \textsc{gpt-4o}\footnote{Preliminary experiments using \textsc{gpt-4o-mini} and \textsc{o3-mini} showed qualitatively similar behavior.} using temperature \texttt{T = 1} for stochasticity. For out-of-sample survey rounds (2024Q1 to 2025Q2), we report accuracy and win-share results only for horizons where realized data is available: current year (CY) and next year (CY+1). For the HICPX variables, only 34 rounds are available as it was only introduced to the survey in 2016Q4.

% specificare che i risultati nel main paper sono tutti stati ottenuti con gpt-4o
% commentare perché ci sono solo 29 surveys per HICPX
% commentare perché per i wins abbiamo dei numeri decimali --> spiegare come sono stati calcolati = media dei win counts ottenuti con ogni persona $\frac{1}{2368} \cdot \sum_{p=1}^{2368} w_{vh}^p$

\subsection{Persona ablation effect}
Our primary methodological contribution examines whether detailed persona descriptions improve forecasting accuracy. 
We conduct a controlled ablation experiment comparing the results of our original prompt against 100 baselines in which we remove the persona description from the prompt. This results in a total of 5,000 forecasts which we can compare to the persona-enhanced ones. 
The results show no statistically significant difference in forecasting performance. A paired t-test on match-level median absolute errors yields a mean difference of 0.01 percentage points (personas minus no-personas), with \(t = -1.02\), \(p = 0.31\). This finding is corroborated by a size-matched Kolmogorov-Smirnov test (\(D = 0.05\), \(p = 0.28\)) showing that error distributions are statistically indistinguishable (see Figure \ref{fig:persona_ablation}).
This null result has significant practical implications: sophisticated persona engineering provides no measurable forecasting advantage and can be omitted to reduce computational costs without sacrificing accuracy. The finding suggests that model performance depends primarily on data quality and task framing rather than prompt elaboration.

\subsection{Panel disagreement}
The dispersion of forecasts within each panel reveals a significant difference between the AI and human panels of forecasters, as shown in Table \ref{tab:variance_comparison}. AI personas exhibit near-zero disagreement, with median inter-quantile ranges (IQRs) mostly below 0.001 percentage points across all variables and horizons, roughly two orders of magnitude lower than human forecasters. Human forecasters, display substantially higher dispersion, with higher median IQRs ranging from 0.17 percentage points (UNR current-year) to slightly higher values at longer time horizons. The disparity is equally pronounced when measured by standard deviation, with AI personas exhibiting roughly one order of magnitude less variation than human forecasters across all variables and horizons. Both dispersion measures confirm that despite the variety of persona prompts, the model converges on mostly homogeneous forecasts, suggesting limited sensitivity to prompt variations in this task domain.

\subsection{Point forecast accuracy}
The mean absolute error results, reported in Table \ref{tab:forecast_accuracy}, show that AI and human forecasters often perform at remarkably similar levels, with identical errors observed in seven of sixteen in-sample comparisons and numerous cases where differences are minimal (within 0.05-0.10 percentage points). The largest performance gaps emerge in specific variable-horizon combinations: AI substantially outperforms humans for GDP growth at the CY+2 horizon (0.60 vs 0.90) and unemployment at current-year out-of-sample forecasts (0.05 vs 0.15), while humans show clear advantages for unemployment at current-year in-sample (0.10 vs 0.20) and HICP current-year out-of-sample (0.01 vs 0.10). The transition from in-sample to out-of-sample periods shows no systematic performance degradation. Despite the model forecasting economic conditions entirely absent from its training data (October 2023 cutoff), accuracy levels remain broadly comparable to the in-sample period, suggesting the model effectively utilizes the real-time economic context provided in prompts rather than relying purely on memorized patterns.

\subsection{Win-share analysis}
In addition to evaluating point accuracy, we compute win-share scores to compare AI and human performance net of ties. The aggregated results are shown in Table \ref{tab:win_share}, both for in-sample and out-of-sample rounds. Appendix D additionally reports the win shares for each survey round by horizon and variable.
The results demonstrate statistically significant yet practically modest differences in forecasting accuracy, with performance patterns varying systematically across variables and horizons. Despite uniform statistical significance at the 1\% level, many win rate differentials are relatively narrow—particularly for inflation forecasts where margins often fall within 1-5 percentage points. The data reveal variable-specific comparative advantages: AI consistently outperforms on core inflation (HICPX) across most horizons, while humans maintain advantages in short-term GDP and unemployment forecasting that gradually erode at longer horizons. The out-of-sample results show more unstable results, with some outcome reversals compared to the in-sample. The limited out-of-sample observations (N=4-6) make it difficult to determine whether these reversals reflect genuine performance differences, structural breaks in the post-2021 period, or simply small-sample volatility.
%However, the absence of systematic degradation provides evidence that \textsc{gpt-4o} can generalize beyond its training data when provided with appropriate real-time economic context.

% \input{tables/T4A-winshare}
% \input{tables/T4B-winshare}

\section{Future work}\label{sec:discussion}
%\TODO{write up discussion}
%Our persona ablation revealed that these prompts components contribute negligibly to forecasting accuracy. A natural extension would systematically ablate other prompt components to identify which contextual information genuinely improves model performance in this task. Specifically, further research should test the contribution of: (1) ECB monetary policy communications (press releases and statements), (2) past SPF median forecasts, and (3) real-time macroeconomic data snapshots. Given that our persona experiment showed elaborate prompt engineering can be ineffective, understanding which economic context actually drives forecasting accuracy would have significant practical implications for prompt design in financial applications.
Having established that persona descriptions are expendable, future work should systematically ablate other prompt components—ECB policy communications, past SPF medians, and real-time macro data—to identify which contextual information genuinely drives forecasting performance versus merely increasing token costs.
Several other extensions merit investigation. First, evaluating density forecasts---which are included in the ECB SPF---rather than just point estimates would test whether LLMs can meaningfully quantify uncertainty. Second, alternative prompting strategies beyond personas, such as explicit chain-of-thought reasoning or adversarial perspectives, may prove more effective at generating forecast diversity. Finally, extending the out-of-sample evaluation beyond our limited six rounds would provide more robust evidence of generalization.

\section{Conclusion}\label{sec:conclusion}
We present the first systematic replication of the ECB Survey of Professional Forecasters using LLMs, evaluating over 2,000 synthetic personas extracted from the PersonaHub corpus across 50 quarterly rounds.
Our controlled ablation experiment reveals that adding these descriptions to the prompt provides no measurable forecasting advantage, with statistical tests showing no significant difference between persona-enhanced and baseline approaches. However, we find that LLMs can achieve competitive accuracy with human forecasters, even on out-of-sample data from 2024-2025 that was entirely absent from model training.
These results have practical implications for AI-assisted forecasting systems. Rather than investing computational resources in elaborate persona engineering, practitioners should focus on robust data integration and model improvements. Our findings also reveal behavioral differences between AI and human forecasting panels: despite diverse prompting, LLMs exhibit very low dispersion and consensus-seeking behavior, in contrast with the heterogeneity observed in human expert panels.
Future research should explore density forecasting capabilities and scenario coherence across multiple variables, while investigating whether alternative prompt engineering approaches beyond persona descriptions can enhance LLM forecasting performance in economic applications.
\bibliographystyle{ACM-Reference-Format}
\bibliography{bio, bio2}
\clearpage
\appendix

\section{Zero-shot Relevance Rating Prompt}
\label{app:relevancerating}

This appendix presents the prompt used to evaluate personas according to three criteria: \textit{EU-centrality}, \textit{neutrality}, and \textit{monetary policy depth}.

\subsection{System and Task Instructions}
\begin{lstlisting}[breaklines=true, basicstyle=\small\ttfamily]
You are assessing expert personas for their suitability in euro-area monetary-policy research.
Return one JSON object only-no additional text.

TASK
- Read the biography supplied by the user.
- Evaluate it against the three pass-fail criteria below.
- Provide a concise one-sentence reason for each decision.
\end{lstlisting}

\subsection{Relevance Criteria}
\begin{lstlisting}[breaklines=true, basicstyle=\small\ttfamily]
CRITERIA
1. Euro-area centrality 
   Fail: Focus is non-EU or purely global with no euro-area anchor.
   Pass: The euro area or an ECB institution is mentioned
       - this includes references to EU countries, central banks or Europe in general,
       - references to other contexts are allowed as long as the euro-area context is mentioned.  
2. Monetary-policy depth
   Fail: Monetary policy is not mentioned at all, or only mentioned in passing with none of the above signals present.
   Pass: The biography engages substantively with monetary policy by satisfying at least one of:
       - names an operational tool (e.g., deposit rate, APP/PEPP, LSAP),
        - discusses a recognised policy rule or doctrine (e.g., Taylor rule, money-growth targeting, rules vs. discretion),
       - analyses a transmission channel or macro outcome (inflation, output, employment, exchange rate, asset prices),
       - references an empirical method used to evaluate policy (event study, VAR, DSGE, natural experiment).
3. Neutrality
    Fail: The biography expresses opinion, advocacy or bias. Look for:  
        - Emotive or value-laden terms ("reckless", "dangerous", "unsustainable").  
        - Framing of personal advocacy or judgment ("skeptical of...", "a critic of...", "optimistic about..."). 
        - Any implicit stance that goes beyond analysis.
    Pass: Tone is descriptive, analytical, or exploratory, without any judgment, prescription, or stance.
\end{lstlisting}

\subsection{Output Schema}
\begin{lstlisting}[breaklines=true, basicstyle=\small\ttfamily]
{
  "euro_area_centrality": "pass" | "fail",
  "monetary_policy_depth": "pass" | "fail",
  "neutrality": "pass" | "fail",
  "notes": {
    "euro_area_centrality": "<one-sentence reason>",
    "monetary_policy_depth": "<one-sentence reason>",
    "neutrality":  "<one-sentence reason>"
  }
}
\end{lstlisting}

\subsection{User Biography}
\begin{lstlisting}[breaklines=true, basicstyle=\small\ttfamily]
USER BIOGRAPHY
 
A researcher who studies the impact of unemployment on time allocation and its implications for household consumption, particularly focusing on the role of unpaid work in the economy. This individual is likely to be interested in understanding how the Great Recession affected people's time allocation patterns and the potential economic costs of involuntary unemployment.
\end{lstlisting}

\section{Persona Examples from PersonaHub Dataset}
\label{app:personas}

This appendix provides examples of the economics-related persona descriptions contained in the PersonaHub dataset, evaluated on the three dimensions relevant to our study: EU-centrality, neutrality, and monetary policy expertise. Only personas meeting all three criteria were retained for the experiments.

\subsection{Retained Persona Example}

\textbf{Financial Economist (EU-focused):} A financial economist who specializes in the analysis of economic cycles and monetary policy. This persona is interested in areas such as the synchronisation of the euro area's economic cycle with that of the US, and how this affects the implementation of monetary policies. They are also interested in the factors that contribute to the degree of synchronisation and how they differ between the euro area and the US. This persona meets our EU-centrality criterion through explicit focus on euro area dynamics, maintains neutrality by presenting balanced analytical perspectives, and demonstrates clear monetary policy expertise.

\subsection{Excluded Persona Examples}

\textbf{Global Economist (US-biased):} A global economist with America Merrill Lynch, with expertise in inflation and deflation, particularly in the context of the U.S. and Europe. They are optimistic about the potential for economic growth but express caution about the risks of shocks that could trigger deflation. They are concerned about the risks of a European deflation, and the potential for a global financial crisis. This persona was excluded due to insufficient EU-centrality (primary focus on US markets) and non-neutral stance (explicit optimism bias).

\textbf{Technology Skeptic (Non-expert):} A technologist who is skeptical of the effectiveness of information technology in stimulating economic growth. This persona believes that technology must be implemented and funded in order to create economic growth. They also believe that the central bank's role in manipulating financial markets is a major impediment to economic growth. This persona is interested in the impact of new technologies on economic growth and the role of savings in promoting capital formation. This persona was excluded for lacking EU-centrality, exhibiting strong ideological bias against central bank intervention, and having insufficient monetary policy expertise.ing economic growth. This persona believes that technology must be implemented and funded in order to create economic growth. They also believe that the central bank's role in manipulating financial markets is a major impediment to economic growth. This persona is interested in the impact of new technologies on economic growth and the role of savings in promoting capital formation. This persona was excluded for lacking EU-centrality, exhibiting strong ideological bias against central bank intervention, and having insufficient monetary policy expertise.

%CAMBIAMO IL NOME DA PROMPT A UN NOME SPECIFICO? perché ora abbiamo il prompt per il relevance rating anche
\section{Prompt}\label{app:prompt}

This appendix contains the complete prompt used to simulate responses to the European Central Bank's Survey of Professional Forecasters (ECB-SPF).

\subsection{System Instructions and Persona Description}

\begin{lstlisting}[breaklines=true, basicstyle=\small\ttfamily]
SYSTEM:
You are participating in the European Central Bank's Survey of Professional Forecasters (ECB-SPF) for round: 2013Q1.
Today's date is 16/01/2013.
You will be asked to provide point forecasts for a set of key macroeconomic indicators (inflation, core inflation, GDP growth and unemployment) for the euro area at different time horizons.

You are: A political economist who specializes in the study of macroeconomics and international trade. Their research interests include the impact of monetary and fiscal policies on economic growth, the effects of currency devaluations on trade deficits, and the role of international capital flows in shaping economic development. They are particularly interested in the implications of the Eurozone for the economies of the PIIGS countries, the effects of financial crises on economic recovery, and the use of monetary and financial restructuring to prevent default. They have experience in analyzing the impacts of trade deficits, currency devaluations, and financial crises on economic growth and development, and have a strong interest in understanding the complex interplay between domestic and international factors in shaping economic outcomes.
\end{lstlisting}

\subsection{Economic Context and Data}

\begin{lstlisting}[breaklines=true, basicstyle=\small\ttfamily]
ECONOMIC CONTEXT:
- Latest realized data:
Variable  Period  Value
HICP      2012Dec  2.20
HICPX     2012Dec  1.50
UNR       2012Nov  12.00
RGDP      2012Q3   -1.00

- Median forecasts from the previous SPF round:
Variable  Horizon  Median forecast
HICP      2013     1.80
HICP      2014     1.80
HICP      2015     1.90
HICP      2017     2.00
HICPX     2013     
HICPX     2014     
HICPX     2015     
HICPX     2017     
UNR       2013     12.10
UNR       2014     11.90
UNR       2015     11.40
UNR       2017     9.40
RGDP      2013     -0.10
RGDP      2014     1.02
RGDP      2015     1.50
RGDP      2017     1.70

- ECB monetary policy communication from the latest Governing Council meeting on 2013-01-10:
Ladies and gentlemen, the Vice-President and I are very pleased to welcome you to our press conference.

Let me wish you all a Happy New Year. We will now report on the outcome of today's meeting of the Governing Council.

Based on our regular economic and monetary analyses, we decided to keep the key ECB interest rates unchanged. HICP inflation rates have declined over recent months, as anticipated, and are expected to fall below 2% this year. Over the policy-relevant horizon, inflationary pressures should remain contained. The underlying pace of monetary expansion continues to be subdued. Inflation expectations for the euro area remain firmly anchored in line with our aim of maintaining inflation rates below, but close to, 2% over the medium term. The economic weakness in the euro area is expected to extend into 2013. In particular, necessary balance sheet adjustments in financial and non-financial sectors and persistent uncertainty will continue to weigh on economic activity. Later in 2013 economic activity should gradually recover. In particular, our accommodative monetary policy stance, together with significantly improved financial market confidence and reduced fragmentation, should work its way through to the economy, and global demand should strengthen. In order to sustain confidence, it is essential for governments to reduce further both fiscal and structural imbalances and to proceed with financial sector restructuring.
\end{lstlisting}

\noindent \textit{[The complete ECB communication continues for several additional paragraphs and has been truncated here for space. The full text was included in the actual prompt.]}

\subsection{Task Instructions and Output Requirements}

\begin{lstlisting}[breaklines=true, basicstyle=\small\ttfamily]
TASK:
You are asked to provide one **numeric point forecast** for each target macroeconomic variable listed below, at multiple time horizons. Do not use ranges or confidence intervals.
All forecasts should be expressed in percent (%), do not include units in the answer.

TARGETS:
For each of the following variables, provide a point forecast:

- HICP: HICP inflation
- HICPX: HICP inflation excluding food and energy
- RGDP: Real GDP growth
- UNR: Unemployment rate

Each variable should be forecast at the following horizons:
- t0: current calendar year (2013)
- t1: next year (2014)
- t2: year after next (2015)
- lt: long-term (2017)

OUTPUT SCHEMA:
Respond strictly in the following JSON format:
{
  "forecasts": [
    {
      "variable": "hicp",
      "horizon": "t0",
      "value": "<<numeric>>"
    },
    {
      "variable": "hicp",
      "horizon": "t1",
      "value": "<<numeric>>"
    },
    ...
    {
      "variable": "unr",
      "horizon": "lt",
      "value": "<<numeric>>"
    }
  ]
}

Reply only with the JSON, no additional text.
\end{lstlisting}

\section{Win-share Analysis}\label{app:win-share}
This appendix presents heatmaps comparing the forecast accuracy of AI and human experts for each target variable (HICP, HICPX, rGDP, UNR) across different forecast horizons (t0, t1, t2, lt) and survey rounds. A win rate of 0\% indicates that human forecasts strictly outperformed AI 100\% of the time, while a win rate of 100\% means AI forecasts were consistently better. Ties are excluded from the win share calculation. The figures reveal that, with some exceptions (for example, the HICP inflation rate between 2012 Q4 and 2015 Q1), AI does not consistently achieve higher win shares compared to human experts. Notably, for inflation forecasts over the long-term horizon, the results indicate that AI's performance is not better than that of human forecasters, as AI's win shares are nearly all 0\%.

% non penso ci sia bisogno di spiegare il perché non ci siano tutti i surveys per ogni horizon --> dovremmo averlo detto già prima.

\onecolumn

\subsection{HICP Inflation Rate}
\begin{figure}[htbp]
  \centering
  \includegraphics[width=0.7\textwidth]{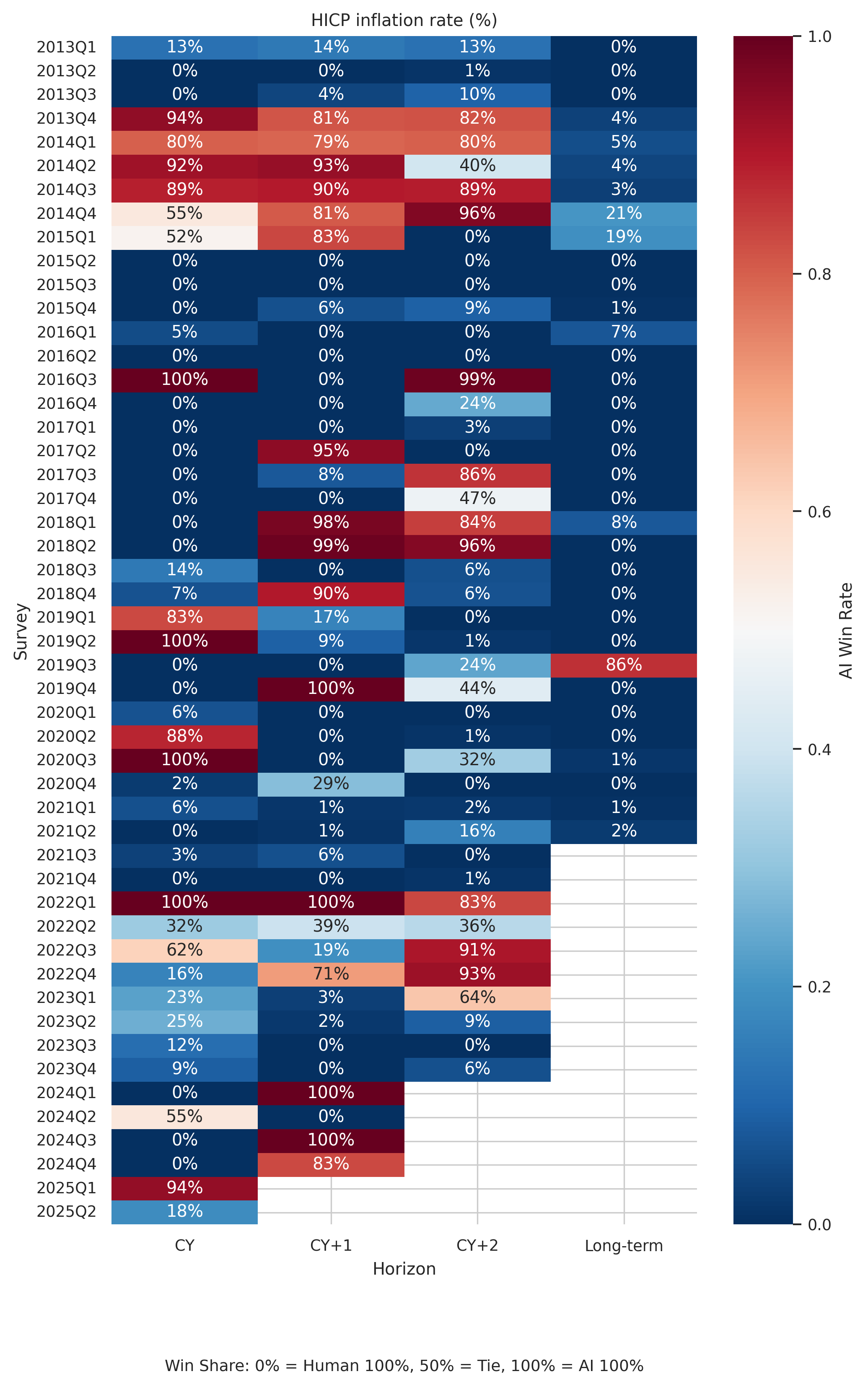}
  \Description[HICP heatmap: AI vs human forecasts]{Heatmap showing the percentage of times AI forecasts for HICP inflation were strictly more accurate than human forecasts across various survey rounds and forecast horizons. Ties are excluded.}
  \caption{AI win share for HICP inflation rate forecasts across horizons and survey rounds.}
\end{figure}

\subsection{HICP Core Inflation Rate}
\begin{figure}[htbp]
  \centering
  \includegraphics[width=0.7\textwidth]{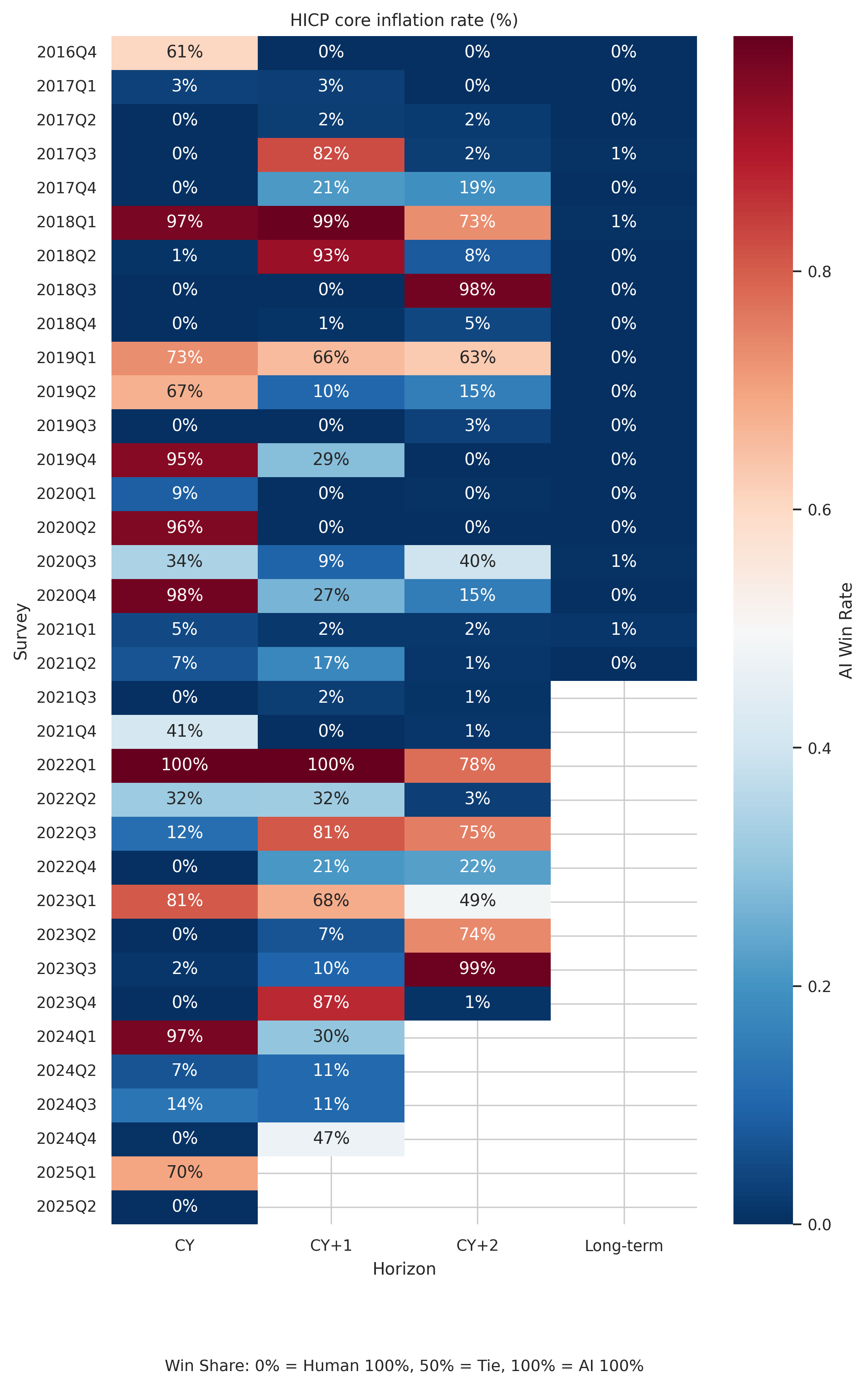}
  \Description[HICPX heatmap: AI vs human forecasts]{Heatmap showing the percentage of times AI forecasts for core HICP inflation (HICPX) were strictly more accurate than human forecasts across survey rounds and forecast horizons. Ties are excluded.}
  \caption{AI win share for HICPX core inflation rate forecasts across horizons and survey rounds.}
\end{figure}

\subsection{Real GDP Rate}
\begin{figure}[htbp]
  \centering
  \includegraphics[width=0.7\textwidth]{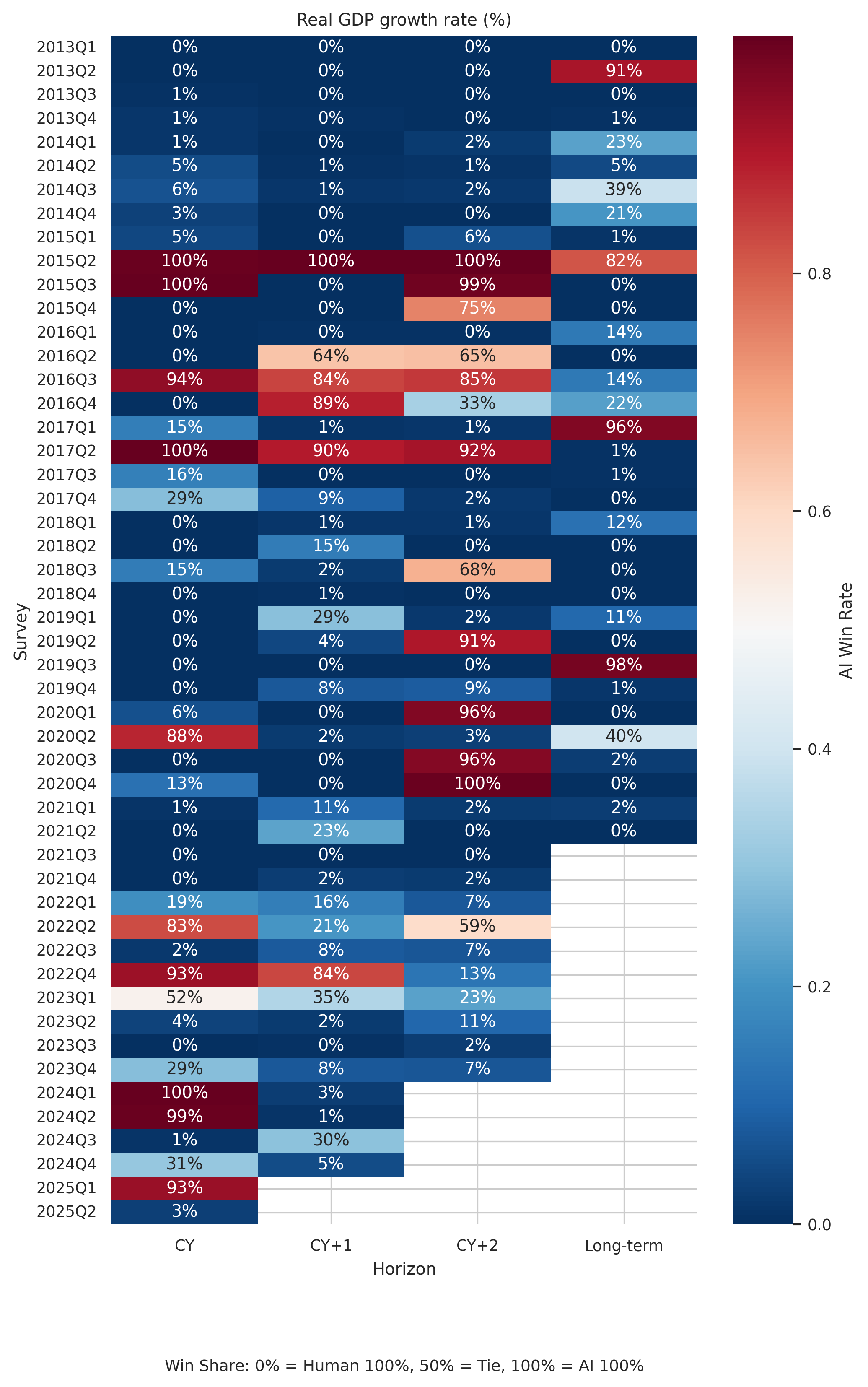}
  \Description[Real GDP heatmap: AI vs human forecasts]{Heatmap showing the percentage of times AI forecasts for real GDP growth were strictly more accurate than human forecasts across various survey rounds and horizons. Ties are excluded.}
  \caption{AI win share for real GDP growth forecasts across horizons and survey rounds.}
\end{figure}

\subsection{Unemployment Rate}
\begin{figure}[htbp]
  \centering
  \includegraphics[width=0.7\textwidth]{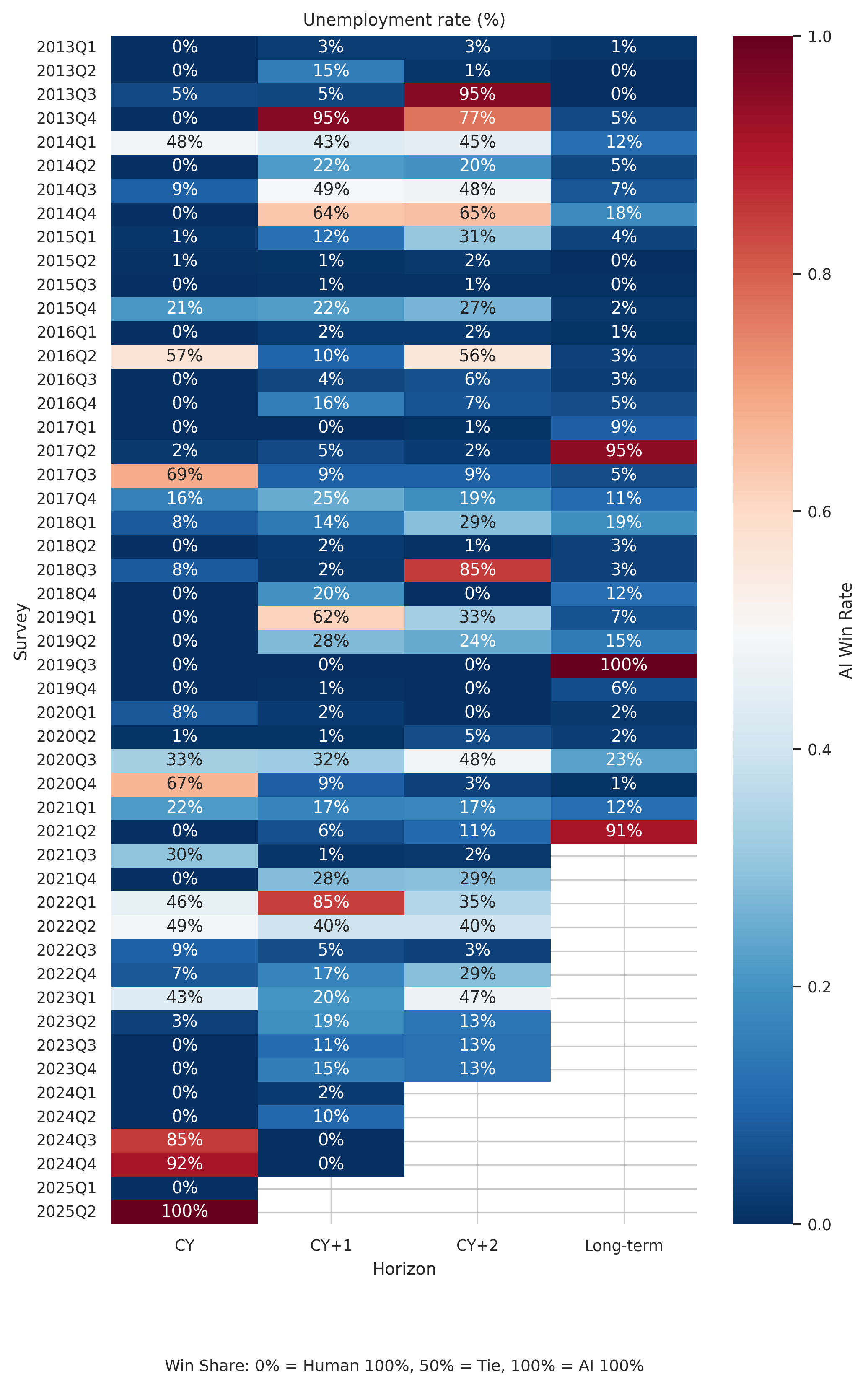}
  \Description[Unemployment heatmap: AI vs human forecasts]{Heatmap showing the percentage of times AI forecasts for the unemployment rate were strictly more accurate than human forecasts across survey rounds and forecast horizons. Ties are excluded.}
  \caption{AI win share for unemployment rate forecasts across horizons and survey rounds.}
\end{figure}

\twocolumn

%add descriptions to images
\end{document}